\documentclass[sigconf]{acmart}
\usepackage {fancyhdr} 
\renewcommand\footnotetextcopyrightpermission[1]{} 

\usepackage{xcolor}
\usepackage{forest}
\usepackage{tikz}
\usepackage{multirow}
\usepackage{subcaption}
\usetikzlibrary{trees}
\definecolor{hidden-draw}{rgb}{0.5, 0.5, 0.5}
\definecolor{harvestgold}{rgb}{0.85, 0.57, 0.0}
\definecolor{cyan}{rgb}{0.0, 1.0, 1.0}
\definecolor{lightcoral}{rgb}{0.94, 0.5, 0.5}
\definecolor{SlateBlue}{rgb}{0.416, 0.353, 0.804}
\definecolor{LightGreen}{rgb}{0.564, 0.933, 0.564}
\definecolor{Turquoise}{rgb}{0.251, 0.878, 0.816}
\definecolor{BlueGreen}{rgb}{0.643, 0.859, 0.871}
\definecolor{BlueViolet}{rgb}{0.54, 0.17, 0.89}
\definecolor{RAG}{rgb}{0.639, 0.835, 0}
\definecolor{0}{rgb}{0.92, 0.3, 0.26}
\definecolor{1}{rgb}{0.961, 0.851, 0.471}
\definecolor{1_1}{rgb}{1.00, 0.933, 0.698}
\definecolor{2}{rgb}{0.522, 0.784, 0.953}
\definecolor{2_1}{rgb}{0.722, 0.592, 0.871}
\definecolor{2_2_2}{rgb}{0.561, 0.659, 0.843}
\definecolor{3}{rgb}{0.56, 0.93, 0.56}
\definecolor{3_1_1}{rgb}{0.808, 0.996, 0.808}
\definecolor{4}{rgb}{0.988, 0.541, 0.565}
\definecolor{4_1}{rgb}{0.996, 0.796, 0.808}
\definecolor{4_1_1}{rgb}{0.957, 0.847, 0.843}
\definecolor{bounding}{rgb}{0.855, 0.392, 0.357}
\definecolor{6}{rgb}{0.690, 0.612, 0.522}

\AtBeginDocument{%
  }

\begin{document}

\title{A Survey on Data Synthesis and Augmentation for Large Language Models}

\author{Ke Wang}
\email{onecall@buaa.edu.cn}
\affiliation{%
  \institution{Hangzhou Innovation Institute, Beihang University}
  \city{}
  \country{}}

\author{Jiahui Zhu}
\email{zhujh224@buaa.edu.cn}
\affiliation{%
  \institution{Hangzhou Innovation Institute, Beihang University}
  \city{}
  \country{}}

\author{Minjie Ren}
\email{rmj_rmj@buaa.edu.cn}
\affiliation{%
  \institution{Hangzhou Innovation Institute, Beihang University}
  \city{}
  \country{}}

\author{Zeming Liu}
\email{zmliu@buaa.edu.cn}
\affiliation{%
  \institution{State Key Laboratory of Virtual Reality Technology and Systems, Beihang University}
  \city{}
  \country{}}

\author{Shiwei Li}
\email{shiweili93@buaa.edu.cn}
\affiliation{%
  \institution{Hangzhou Innovation Institute, Beihang University}
  \city{}
  \country{}}

\author{Zongye Zhang}
\email{zhangzongye@buaa.edu.cn}
\affiliation{%
  \institution{State Key Laboratory of Virtual Reality Technology and Systems, Beihang University}
  \city{}
  \country{}}

\author{Chenkai Zhang}
\email{zhangchenkai@buaa.edu.cn}
\affiliation{%
  \institution{State Key Laboratory of Virtual Reality Technology and Systems, Beihang University}
  \city{}
  \country{}}

\author{Xiaoyu Wu}
\email{zf2306113@buaa.edu.cn}
\affiliation{%
  \institution{Hangzhou Innovation Institute, Beihang University}
  \city{}
  \country{}}

\author{Qiqi Zhan}
\email{zhanqiqi@buaa.edu.cn}
\affiliation{%
  \institution{State Key Laboratory of Virtual Reality Technology and Systems, Beihang University}
  \city{}
  \country{}}

\author{Qingjie Liu}
\email{qingjie.liu@buaa.edu.cn}
\affiliation{%
  \institution{State Key Laboratory of Virtual Reality Technology and Systems, Beihang University}
  \city{}
  \country{}}

\author{Yunhong Wang}
\email{yhwang@buaa.edu.cn}
\affiliation{%
  \institution{State Key Laboratory of Virtual Reality Technology and Systems, Beihang University}
  \city{}
  \country{}}

\renewcommand{\shortauthors}{Wang et al.}

\begin{abstract}
The success of Large Language Models (LLMs) is inherently linked to the availability of vast, diverse, and high-quality data for training and evaluation. However, the growth rate of high-quality data is significantly outpaced by the expansion of training datasets, leading to a looming data exhaustion crisis. This underscores the urgent need to enhance data efficiency and explore new data sources. In this context, synthetic data has emerged as a promising solution. Currently, data generation primarily consists of two major approaches: data augmentation and synthesis. This paper comprehensively reviews and summarizes data generation techniques throughout the lifecycle of LLMs, including data preparation, pre-training, fine-tuning, instruction-tuning, preference alignment, and applications. Furthermore, We discuss the current constraints faced by these methods and investigate potential pathways for future development and research. Our aspiration is to equip researchers with a clear understanding of these methodologies, enabling them to swiftly identify appropriate data generation strategies in the construction of LLMs, while providing valuable insights for future exploration. 

\end{abstract}

\maketitle
\section{Introduction}
In recent years, large language models (LLMs) have demonstrated unparalleled capabilities across a wide array of tasks \cite{guo2023close, bang2023multitask, rathje2024gpt}, firmly establishing themselves as the backbone of general artificial intelligence (AI) systems. These models achieve significant improvements in natural language processing \cite{yang2023bigtranslate, zhang2024sentiment, zhang2024benchmarking}, computer vision \cite{wang2024visionllm, yuan2021florence, lai2024lisa}, and other research fields \cite{qiu2023large, yang2023fingpt, cui2023chatlaw}, consistently pushing the boundaries of what AI can achieve. The success of LLMs is largely attributed to their ability to capture intricate patterns and relationships within vast amounts of data, allowing them to perform complex tasks such as natural language inference \cite{liu2024large, dalal2024inference}, visual question answering \cite{manas2024improving, ozdemir2024enhancing}, and vision-and-language navigation \cite{lin2024navcot, schumann2024velma} with remarkable proficiency.
\begin{figure*}[t]
  \centering
  \includegraphics[width=0.87\linewidth]{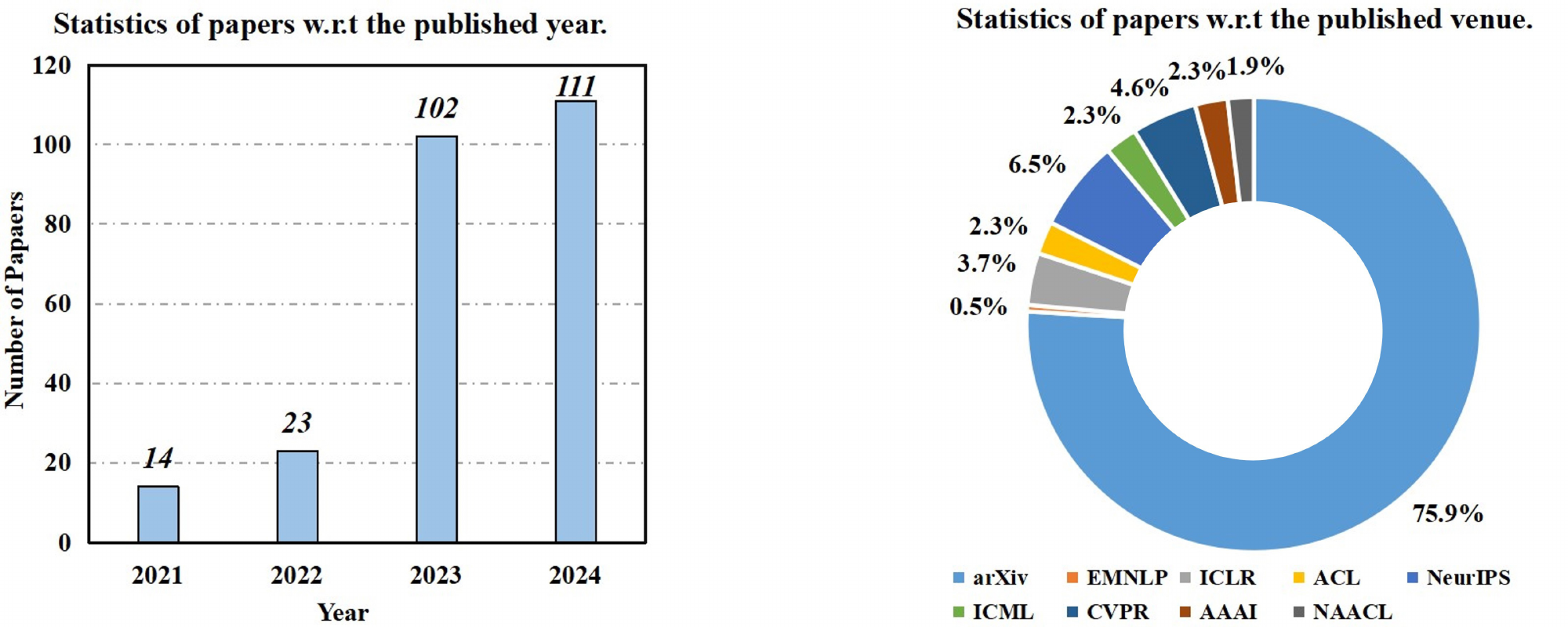}
  \caption{Statistics of the publications related to LLM-oriented data synthesis and augmentation technologies, grouped by the publication year and venue.}
  \label{trend}
\end{figure*}

However, the performance of LLMs is highly dependent on the quality and volume of the data they are trained on \cite{gandhi2024better, achiam2023gpt, floridi2020gpt}. With the exponential growth in model size — now reaching billions or even trillions of parameters \cite{zhao2023survey, le2023bloom, ren2023pangu} — there is an increasing demand for large-scale, diverse, and high-quality data to ensure robust generalization across various tasks and domains. Obtaining such data poses significant challenges due to the high costs of data collection and the problems introduced by privacy concerns. Additionally, the growth rate of high-quality data lags far behind the rapidly increasing size of training datasets. If this trend continues, the available data will eventually be depleted, implying that without significant improvements in data efficiency or the discovery of new data sources, the growth of LLMs may slow down considerably. Given these impending limitations, data synthesis and augmentation techniques become essential to extending the lifespan and generalization of LLMs. Traditional data synthesis and augmentation techniques \cite{takahashi2019data, krell2017rotational, cubuk2019autoaugment, liu2020survey}, such as image rotation, cropping, flipping, and rule-based natural language generation, have been widely used to address these data limitations. Although these approaches improve data diversity and address data scarcity to some extent, they still struggle to fully capture the complexities of real-world data \cite{feng2021survey}, generate data at scale \cite{yang2022image}, and defend against adversarial examples \cite{qiu2020fencebox}, limiting their effectiveness for training LLMs.
\begin{figure*}[t]
    \centering
    \includegraphics[width=0.86\linewidth]{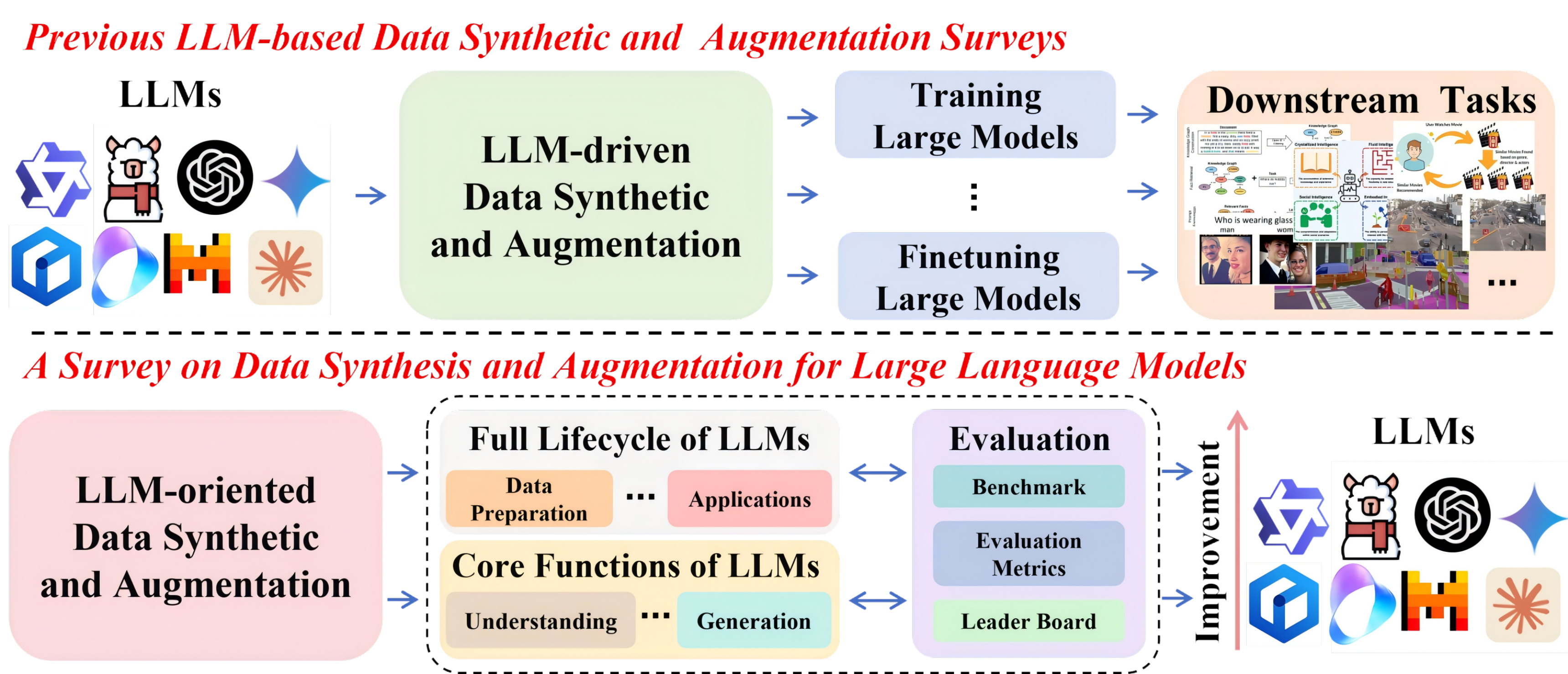}
    \caption{A comparison between existing surveys on data synthesis and augmentation techniques and our work. Previous surveys primarily focus on LLM-based data synthesis and augmentation methods aimed at supporting downstream tasks. In contrast, our work emphasizes LLM-oriented data synthesis and augmentation, systematically covering the full lifecycle of LLMs—from data preparation to applications—and addressing core LLM functions such as understanding and generation, with the ultimate goal of improving LLMs themselves through data-centric techniques.}
    \label{fig:overall}
\end{figure*}

To overcome these challenges, researchers have increasingly turned to LLM-oriented data synthesis and augmentation techniques, recognizing the ability of LLMs to model complex patterns from large datasets and generate synthetic data that closely mirror real-world distributions while introducing valuable variations \cite{zhang2023planning, dai2023chataug, samuel2023can}. These studies reduce the reliance on manually curated datasets and enable the generation of high-quality, diverse data that meets the evolving demands of LLMs throughout their lifecycle and functions. To capture the breadth of these efforts, we collected papers related to LLM-oriented data synthesis and augmentation by searching Google Scholar using keywords such as "data synthesis," "data augmentation," and "large models." Figure \ref{trend} illustrates the publication trends by year and venue, reflecting the increasing interest in this field. As of October 2024, we identified $250$ unique publications covering diverse research topics and venues. Summarizing these efforts provides critical insights into the progress and challenges that remain, offering a foundation for future research. Despite these advancements, several key challenges remain in LLM-oriented data synthesis and augmentation. The misuse of synthetic data poses risks, particularly in spreading misinformation and raising ethical concerns around manipulating public opinion. Additionally, synthetic data often introduces ambiguity when aligning AI models with human values, potentially leading to biased outcomes. Evaluating models trained on synthetic data is also complex, as traditional benchmarks may not fully capture the nuances of this data. Ensuring reliability is another concern, as biases and inaccuracies from original datasets can persist in synthetic data, limiting its generalization across domains. Moreover, the computational demands of LLMs, along with challenges in handling less common languages or novel instructions, complicate broader applications. Finally, the lack of a unified framework for organizing and comparing the methods proposed in both academia and industry remains a barrier for researchers navigating this rapidly evolving field.

This survey aims to address these gaps by providing a comprehensive overview of LLM-oriented data synthesis and augmentation techniques. As shown in Figure \ref{fig:overall}, unlike previous surveys \cite{long2024llms, ma2024investigating, zhou2024survey, wei2023simple, ding2024data}, which primarily focus on applying these methods to support specific downstream tasks or particular stages of LLMs, our work emphasizes the direct role of LLM-oriented techniques in improving the overall performance of LLMs across various stages of their lifecycle and core functions. In contrast to the work \cite{liuBestPracticesLessons2024}, which focuses on practices for synthetic data generation to address challenges like data scarcity and privacy, our survey extends beyond practical guidance by categorizing methods aimed at improving LLM performance holistically. We examine not only data generation but also how these techniques enhance LLMs across all stages and functions, offering a more integrated, data-centric framework for advancing LLMs. Specifically, we systematically review and categorize existing research from two key perspectives: the lifecycle of LLMs (from pre-training to fine-tuning and application) and their core functions (understanding, logic, memory, and generation). By framing the discussion around these dual perspectives, we offer clearer insights into the development, interconnections, and practical applications of different approaches. Moreover, we identify critical challenges, explore emerging research directions, and highlight potential breakthroughs that could further drive advancements in LLM performance through data-centric methods. 

The contributions of this survey are summarized as follows:
\begin{itemize}
  \item \textbf{First Survey:} To our knowledge, we present the first comprehensive survey focused on advancing LLMs through data synthesis and augmentation, systematically covering the entire lifecycle stages and core functions of LLMs. This survey provides an in-depth analysis of current methodologies and highlights the unique challenges at each stage.
  \item \textbf{New taxonomy:} We introduce an innovative organizational framework that categorizes existing research from two key perspectives: the lifecycle stages of LLMs and their core functions. This taxonomy offers a clearer understanding of the progression, interconnections, and applicability of different approaches, providing valuable insights into both developmental and functional aspects of LLM-oriented data synthesis and augmentation.
  \item \textbf{New frontiers:} We identify critical challenges, and explore emerging research directions, and potential breakthroughs in LLM-oriented data synthesis and augmentation. This discussion aims to inspire future research and guide developments in data-centric techniques for LLM advancement.
  \item \textbf{Abundant resources:} We organize and maintain a dedicated repository to support ongoing research and collaboration in LLM-oriented data synthesis and augmentation. This resource includes a curated collection of related papers, multiple leaderboards tracking the latest advancements, and regular updates to foster innovation, guide future research directions, and accelerate breakthroughs in the field.
\end{itemize}

By offering a comprehensive overview of LLM-oriented data synthesis and augmentation approaches, this survey aims to clarify the current state of the field and inspire future research directions that can further enhance LLM capabilities through data synthesis and augmentation methodologies.

We organize the remainder of this survey as follows: Section 2 categorizes the primary areas of LLM-oriented data synthesis and augmentation, providing an overview of the foundational techniques. Section 3 discusses the current LLM-oriented data synthesis and augmentation methods from the perspective of the full lifecycle of LLMs, detailing how these techniques are employed at different stages of model development. In Section 4, we review these methods from the viewpoint of core LLM functions, exploring how data synthesis and augmentation enhance key capabilities such as understanding, logic, memory, and generation. Section 5 delves into the evaluation strategies for LLM-oriented data synthesis and augmentation, addressing benchmarks, evaluation metrics, and leaderboards used to assess and compare the effectiveness of existing approaches. Finally, Section 6 provides insights into challenges and emerging trends in LLM-oriented data synthesis and augmentation, offering recommendations for future research directions that can contribute to the continued advancement of LLMs through data synthesis and augmentation methodologies. 

\begin{figure*}[t]
  \centering
  \resizebox{\textwidth}{0.88\textheight}{
    \begin{forest}
      for tree={
        grow=east,
        reversed=true,
        anchor=base west,
        parent anchor=east,
        child anchor=west,
        base=left,
        font=\normalsize,
        rectangle,
        draw=hidden-draw,
        rounded corners,
        align=left,
        minimum width=4em,
        edge+={darkgray, line width=1pt},
        s sep=3pt,
        inner xsep=2pt,
        inner ysep=3pt,
        edge path={
          \noexpand\path [draw, \forestoption{edge}]
          (!u.parent anchor) -- ++(1.5mm,0) |- (.child anchor) \forestoption{edge label};},
        ver/.style={rotate=90, child anchor=north, parent anchor=south, anchor=center},
      },
      where level=1{text width=7.2em,font=\small,}{},
      where level=2{text width=8.0em,font=\small,}{},
      where level=3{text width=10.0em,font=\small,}{},
      where level=4{text width=9.0em,font=\small,}{},
      where level=5{text width=6.4em,font=\small,}{},
      [
        Data Synthesis and Augmentation for Large Language Models: A Survey, ver, color=bounding!100, fill=0!15, text=black, font=\normalsize, text width=32.0em, text centered
        [
          Taxonomy (\S \ref{sec:background}), color=1!100, fill=1!80, text=black
            [
                Data Augmentation \\ (\S \ref{subsec:data augmentation}), color=1!100, fill=1!65, text=black
                [                    
                    Data Labeling, color=1!100, fill=1_1!55, text=black
                    [
                    {~T-SciQ~\cite{wang2024t}, ChatGPT-based~\cite{zhu2023can,gilardi2023chatgpt,alizadeh2023open}}, color=bounding!70, fill=1_1!20, text=black, text width=24.0em
                    ]
                ]                    
                [
                    Data Reformation, color=1!100, fill=1_1!55, text=black
                    [
                    {~Mosaic\cite{Jocher_YOLOv5_by_Ultralytics_2020}, CORE~\cite{dixit2022core}, ALIA~ \cite{dunlap2023diversify}, ChatAug ~\cite{dai2023chataug}}, color=bounding!70, fill=1_1!20, text=black, text width=24.0em
                    ]                    
                ]
                [
                    Co-Annotation, color=1!100, fill=1_1!55, text=black
                    [
                    {~Co-annotating~\cite{li2023coannotating}, ToolCoder~\cite{zhang2023toolcoder}}, color=bounding!70, fill=1_1!20, text=black, text width=24.0em
                    ]
                ]
            ]
            [
                Data Synthesis (\S \ref{subsec:data synthesis}), color=1!100, fill=1!65, text=black
                [                    
                    General Model \\ Distillation, color=1!100, fill=1_1!55, text=black
                    [
                    {~TinyStories\cite{eldan2023tinystories},~Phi-1\cite{gunasekar2023textbooks,li2023textbooks}, Alpagasus \cite{chen2023alpagasus}, WizardLM \cite{xu2023wizardlm}}, color=bounding!70, fill=1_1!20, text=black, text width=24.0em
                    ]
                ]                    
                [
                    Domain Model \\ Distillation, color=1!100, fill=1_1!55, text=black
                    [
                    {~Minerva\cite{lewkowycz2022solving},~DeepSeek-Prover\cite{xin2024deepseekprover},~WizardCoder\cite{luo2024wizardcoder}}, color=bounding!70, fill=1_1!20, text=black, text width=24.0em
                    ]                    
                ]
                [
                    Model Self-Improvement, color=1!100, fill=1_1!55, text=black
                    [
                    {~Rephrasing\cite{maini2024rephrasing},~Self-instruct\cite{wang2023self}, SPIN \cite{chen2024self}, SelTDA \cite{khan2023q}}, color=bounding!70, fill=1_1!20, text=black, text width=24.0em
                    ]
                ]
            ]
      ]
          [   
            Full Lifecycle \\ of LM (\S \ref{sec:full lifecycle}), color=2!100, fill=2!85, text=black
            [
              Data Preparation \\ (\S \ref{subsec:data preparation}), color=orange!70, fill=orange!35, text=black
              [
                General Model Distillation, color=orange!70, fill=orange!18, text=black
                [
                {~Dialogic~\cite{li2022controllable}, MathInstruct~\cite{yue2023mammoth}, Genixer~\cite{zhao2023genixer}, Magpie~\cite{xu2024magpie}, \\ MMIQC~\cite{liu2024augmenting}, Genie~\cite{yehudai2024genie}, Case2Code~\cite{shao2024case2code}, UltraChat~\cite{ding2023enhancing}}, color=bounding!70, fill=orange!7, text=black, text width=24.0em
                ]
              ]
              [
                Data Augmentation, color=orange!70, fill=orange!18, text=black
                [
                {~Disco~\cite{chen2022disco}, GPT3Mix~\cite{yoo2021gpt3mix}, CoAnnotating~\cite{li2023coannotating}, ALIA~\cite{dunlap2023diversify}, \\FullAnno~\cite{Hao2024FullAnno}, Dialgen~\cite{lu2023dialgen}, TinyGSM~\cite{liu2023tinygsm}, AMPS~\cite{hendrycks2021measuring}}, color=bounding!70, fill=orange!7, text=black, text width=24.0em
                ]
              ]
            ]
            [
              Pretraining (\S \ref{subsec:pre-training}), color=blue!40, fill=blue!20, text=black
              [
                General Model Distillation, color=blue!40, fill=blue!10, text=black
                [
                {~Phi-1~\cite{gunasekar2023textbooks}, SciLitLLM~\cite{li2024scilitllm}, TRAIT~\cite{liang2024task}, AnyGPT~\cite{zhan2024anygpt}, \\ Phi-1.5~\cite{li2023textbooks}, TinyDialogues~\cite{feng2024child}}, color=bounding!70, fill=blue!5, text=black, text width=24.0em
                ]
              ]
              [
                Model Self-Improvement, color=blue!40, fill=blue!10, text=black
                [
                {~VILA-2~\cite{fang2024vila}}, color=bounding!70, fill=blue!5, text=black, text width=24.0em
                ]
              ]
              [
                Data Augmentation, color=blue!40, fill=blue!10, text=black
                [
                {~WRAP~\cite{maini2024rephrasing}, KMLM~\cite{liu2021enhancing}, bioR~\cite{zhu2023physics}, Physics-based~\cite{liu2024large}}, color=bounding!70, fill=blue!5, text=black, text width=24.0em
                ]
              ]
            ]
            [
              Finetuning (\S \ref{subsec:fine-tuning}), color=purple!60, fill=purple!20, text=black
              [
                General Model Distillation, color=purple!60, fill=purple!10, text=black
                [
                {~LAB~\cite{sudalairaj2024lab}, LLM2LLM~\cite{lee2024llm2llm}, GLAN~\cite{li2024synthetic}, Clingen~\cite{xu2024knowledgeinfused}, \\ Baize~\cite{xu2023baize}, Evol-Instruct~\cite{xu2023wizardlm}, HuaTuo~\cite{wang2023huatuo}, NExT-GPT~\cite{wu2023next}}, color=bounding!70, fill=purple!5, text=black, text width=24.0em
                ]
              ]
              [
                Model Self-Improvement, color=purple!60, fill=purple!10, text=black
                [
                {~STaR~\cite{zelikman2022star}, REST~\cite{gulcehre2023reinforced}, Self-Translate~\cite{ri2024self}, Self-Instruct~\cite{wang2023self}, \\ RFT~\cite{yuan2023scaling}, CodeRL~\cite{le2022coderl}, REST-EM~\cite{singh2023beyond}, DeepSeekProver~\cite{xin2024deepseekprover}}, color=bounding!70, fill=purple!5, text=black, text width=24.0em
                ]
              ]
              [
                Data Augmentation, color=purple!60, fill=purple!10, text=black
                [
                {~MathGenie~\cite{lu2024mathgenie}, DISC-MedLLM~\cite{bao2023discmedllm}, MetaMath~\cite{yu2024metamath}, \\ Symbol tuning~\cite{wei2023symbol}, Llama-3-UltraMedical~\cite{zhang2024ultramedical}, Llemma~\cite{azerbayev2023llemma}}, color=bounding!70, fill=purple!5, text=black, text width=24.0em
                ]
              ]
            ]
            [
              Instruction-Tuning\\
               (\S \ref{sec:Instruction-Tuning}),
              color=BlueGreen!150, fill=BlueGreen!90, text=black
              [
                General Model Distillation, color=BlueGreen!120, fill=BlueGreen!50, text=black
                [
                {~Alpaca~\cite{Taori2023alpaca}, Vicuna~\cite{chiang2023vicuna}, Orca~ \cite{mukherjee2023orca},Baize~\cite{xu2023baize}, LLaVA~\cite{liu2024visual}}, color=bounding!70, fill=BlueGreen!30, text=black, text width=24.0em
                ]
              ]
              [
                Model Self-Improvement, color=BlueGreen!120, fill=BlueGreen!50, text=black
                [
                {~Self-Instruct~\cite{wang2023self}, SPIN~\cite{chen2024self}, CAI~\cite{bai2022constitutional}, Toolformer~\cite{schick2024toolformer}}, color=bounding!70, fill=BlueGreen!30, text=black, text width=24.0em
                ]
              ]
              [
                Data Augmentation, color=BlueGreen!120, fill=BlueGreen!50, text=black
                [
                {~T-SciQ~\cite{wang2024t}, CORE~\cite{dixit2022core}, ChatAug ~\cite{dai2023chataug}, ToolCoder~\cite{zhang2023toolcoder}}, color=bounding!70, fill=BlueGreen!30, text=black, text width=24.0em
                ]
              ]
            ]
            [
              Preference \\ Alignment(\S \ref{subsec:preference-alignment}) , color=BlueViolet!60, fill=BlueViolet!25, text=black
              [
                General Model Distillation, color=BlueViolet!60, fill=BlueViolet!15, text=black
                [
                {~ULTRAFEEDBACK~\cite{cui2023ultrafeedback}, HelpSteer~\cite{wang2023helpsteer}, LEMA~\cite{an2023learning}},color=bounding!70, fill=BlueViolet!7, text=black, text width=24.0em
                ]
              ]
              [
                Domain Model Distillation, color=BlueViolet!60, fill=BlueViolet!15, text=black
                [
                {~BAD ~\cite{xu2021bot}, BEAVERTAILS~\cite{ji2024beavertails},  PRM800K~\cite{lightman2023let}, WebGPT~\cite{nakano2021webgpt}},color=bounding!70, fill=BlueViolet!7, text=black, text width=24.0em
                ]
              ]
              [
                Model Self-Improvement, color=BlueViolet!60, fill=BlueViolet!15, text=black
                [
                {~OAIF~\cite{guo2024direct}, SELF-JUDGE~\cite{ye2024self}, SALMON~\cite{sun2024salmon}, SteerLM~\cite{dong2023steerlm}}, color=bounding!70, fill=BlueViolet!7, text=black, text width=24.0em
                ]
              ]
              [
                Data Augmentation, color=BlueViolet!60, fill=BlueViolet!15, text=black
                [
                 {~ Starling-7B~\cite{zhu2023starling}, UltraInteract ~\cite{yuan2024advancing}, CriticBench~\cite{lin2024criticbench}}, color=bounding!70, fill=BlueViolet!7, text=black, text width=24.0em
                ]
              ]
            ]
            [
          Applications (\S \ref{sec:application}), color=red!40, fill=red!25, text=black
          [
            Math, color=4!80, fill=4_1!70, text=black
            [
              {~MetaMath~\cite{yu2024metamath}, MammoTH~\cite{yue2023mammoth}, STaR~\cite{zelikman2022star}, Galactia~\cite{taylor2022galactica}, \\ ~DeepSeekProver~\cite{xin2024deepseekprover}, WizardMath~\cite{luo2023wizardmath}}, color=bounding!70, fill=4_1_1!20, text=black, text width=24.0em
            ]
          ]
          [
            Science, color=4!80, fill=4_1!70, text=black
            [
              {~SciLitLLM~\cite{li2024scilitllm}, ChemLLM~\cite{zhang2024chemllm}, SciGLM~\cite{zhang2024sciglm}, Galactia~\cite{taylor2022galactica}}, color=bounding!70, fill=4_1_1!20, text=black, text width=24.0em
            ]
          ]
          [
            Code, color=4!80, fill=4_1!70, text=black
            [
              {~WizardCoder~\cite{luo2024wizardcoder}, MagicCoder~\cite{wei2024magicoder}, Code Alpaca~\cite{chaudhary2023code}, \\ ~Code LLama~\cite{roziere2024code}, Phi-1~\cite{gunasekar2023textbooks}, Phi-1.5~\cite{li2023textbooks}}, color=bounding!70, fill=4_1_1!20, text=black, text width=24.0em
            ]
          ]
          [
            Medical, color=4!80, fill=4_1!70, text=black
            [
              {~DISC-MedLLM~\cite{bao2023discmedllm}, HuatuoGPT~\cite{zhang2023huatuogpt, chen2023huatuogptii}, ChatCounselor~\cite{liu2023chatcounselor},\\ ~ClinGen~\cite{xu2024knowledgeinfused}}, color=bounding!70, fill=4_1_1!20, text=black, text width=24.0em
            ]
          ]
          [
            Law, color=4!80, fill=4_1!70, text=black
            [
              {~DISC-LawLLM~\cite{yue2023disclawllm}, LawyerLLaMA~\cite{huang2023lawyer}, LawGPT~\cite{zhou2024lawgpt}, \\ ~WisdomInterrogatory~\cite{zhihaillm2024zhihaillm}}, color=bounding!70, fill=4_1_1!20, text=black, text width=24.0em
            ]
          ]
        ]
          ]
          [
            Functionality (\S \ref{sec:function}), color=SlateBlue!80, fill=SlateBlue!40, text=black
            [
              Understanding (\S \ref{sec:Understanding}), color=brown!60, fill=brown!30, text=black
              [
                {~Alpaca~\cite{Taori2023alpaca}, WizardLM \cite{xu2023wizardlm}, WRAP\cite{maini2024rephrasing}, LLaVA \cite{liu2024visual}, ChartLlama\cite{han2023chartllama},Genixer\cite{zhao2023genixer}}, color=bounding!70, fill=brown!10, text=black, text width=35.8em
              ]
            ]
            [
              Logic (\S \ref{sec:Logic}), color=2_2_2!100, fill=2_2_2!60, text=black
              [
                {~ReST$^{EM}$ \cite{singh2023beyond}, Case2Code\cite{shao2024case2code}, MathInstruct\cite{yue2023mammoth}, MMIQC\cite{liu2024augmenting}, STaR\cite{zelikman2022star},SelTDA \cite{khan2023q}}, color=bounding!70, fill=2_2_2!20, text=black, text width=35.8em
              ]
            ]
            [
              Memory (\S \ref{sec:Memory}), color=LightGreen!100, fill=LightGreen!40, text=black
              [
                {~Quiet-STaR \cite{zelikman2024quiet}, AutoKG \cite{zhu2024llms}, Persona Hub \cite{chan2024scaling}, AceCoder \cite{li2023acecoder}, RepoCoder \cite{zhang2023repocoder}}, color=bounding!70, fill=LightGreen!10, text=black, text width=35.8em
              ]
            ]
            [
              Generation (\S \ref{sec:Generation}), color=Turquoise!80, fill=Turquoise!30, text=black
              [
                {~Genie\cite{yehudai2024genie}, UltraMedical\cite{zhang2024ultramedical}, HuaTuo\cite{wang2023huatuo}, TinyStories\cite{eldan2023tinystories}, DIALOGIC\cite{li2022controllable}, ALIA \cite{dunlap2023diversify}}, color=bounding!70, fill=Turquoise!10, text=black, text width=35.8em
              ]
            ]
          ]
        [
          Challenges and \\ Limitations (\S \ref{sec:challenge}), color=6!100, fill=6!50, text=black
          [
            Synthesizing and \\ Augmenting  Method \\ (\S \ref{sec:Synthesizing and Augmenting  Method}), color=6!100, fill=6!35, text=black
            [
              {~d-RLAIF\cite{lee2023rlaif}, LLM2LLM\cite{lee2024llm2llm}, Wizardmath\cite{luo2023wizardmath}, STaR\cite{zelikman2022star}, SciGLM\cite{zhang2024sciglm}, ChemLLM\cite{zhang2024chemllm}
              }, color=bounding!70, fill=6!20, text=black, text width=35.8em
            ]
          ]
          [
            Data Quality (\S \ref{sec:Data Quality}), color=6!100, fill=6!35, text=black
            [
              {~ LLMs4Synthesis\cite{giglou2024llms4synthesis}, CoRAL\cite{wu2024coral}, FORD\cite{xiong2023examining},LTGC\cite{zhao2024ltgc}}, color=bounding!70, fill=6!20, text=black, text width=35.8em
            ]
          ]
          [
            Impact of Data \\ Synthesis and \\ Augmentation (\S \ref{sec:Impact of Data Synthesis and Augmentation}), color=6!100, fill=6!35, text=black
            [
              {~ DataDreamer\cite{patel2024datadreamer},HARMONIC\cite{wang2024harmonic}}, color=bounding!70, fill=6!20, text=black, text width=35.8em
            ]
          ]
          [
            Impact on Different \\ Applications and \\Tasks (\S \ref{sec:Impact on Different Applications and Tasks}), color=6!100, fill=6!35, text=black
            [
              {~PANDA\cite{liu2024panda},REGA\cite{wang2024role}}, color=bounding!70, fill=6!20, text=black, text width=35.8em
            ]
          ]
          [
            Future Directions \\ (\S \ref{sec:Future Directions}), color=6!100, fill=6!35, text=black
            [
              {~TabSynDex\cite{chundawat2022universal},CoLa-Diff\cite{jiang2023cola},WizardCoder\cite{luo2024wizardcoder}, WebGPT\cite{nakano2021webgpt}}, color=bounding!70, fill=6!20, text=black, text width=35.8em
            ]
          ]
        ]
      ]
    \end{forest}
    }
  \caption{The main content flow and categorization of this survey.}
  \label{figure:categorization_of_survey}
\end{figure*}
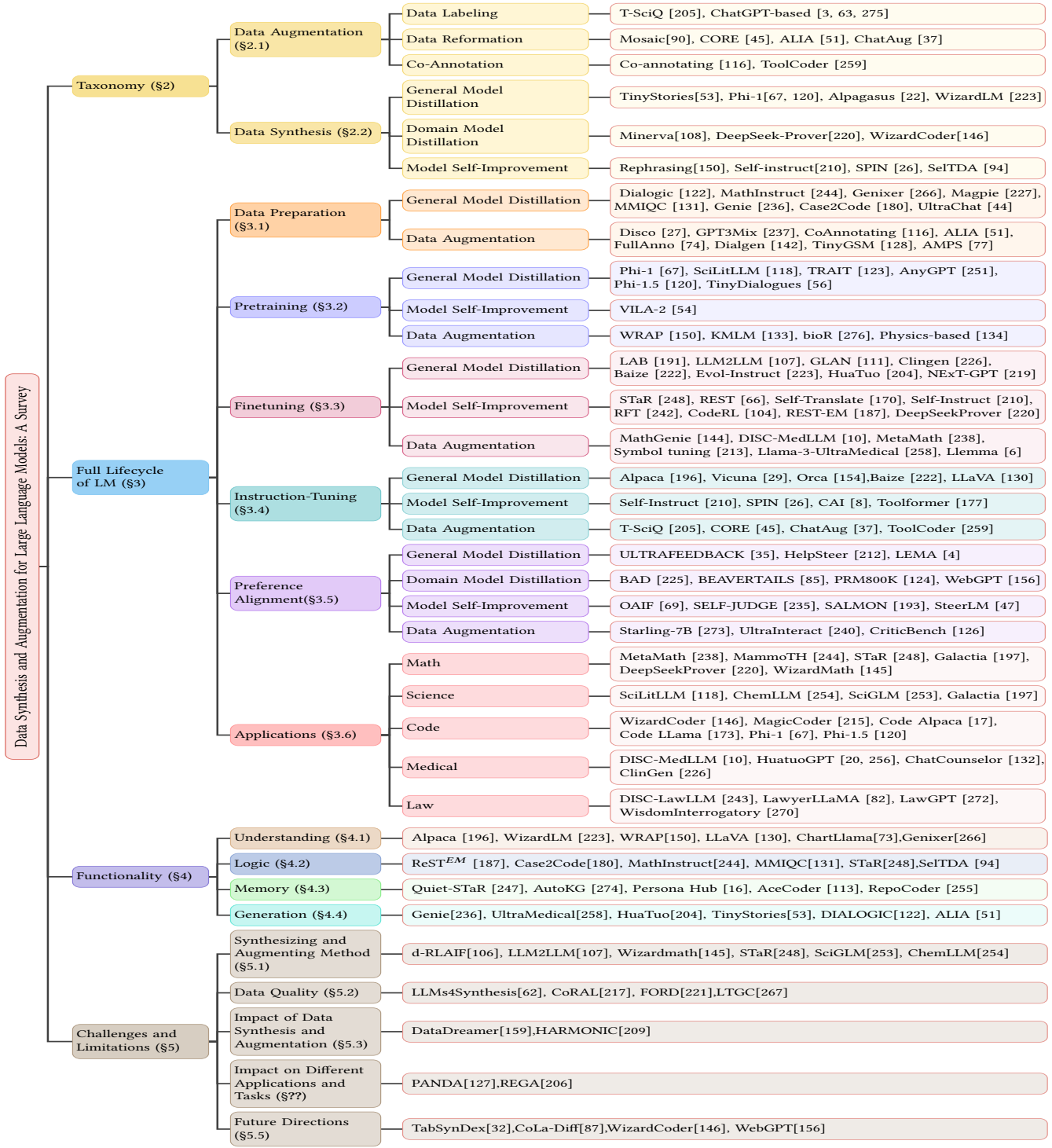

\section{Taxonomy} 
\label{sec:background}
\begin{figure*}[t]
  \centering
  \includegraphics[width=1.0\linewidth]{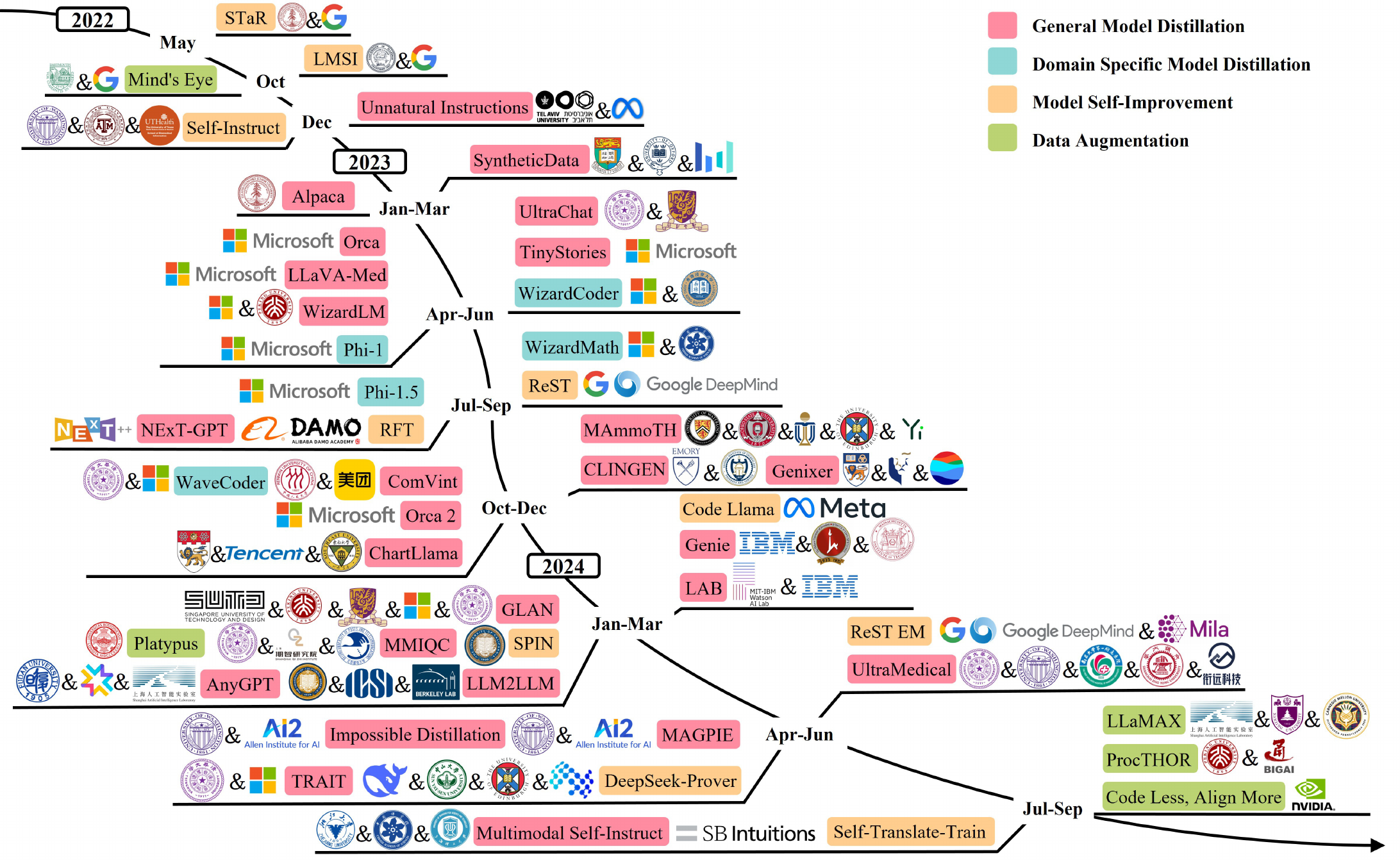}
  \caption{Illustration of the evolutionary steps in the development of data synthesis and augmentation techniques for large models.}
  \label{summary}
\end{figure*}

Data generation methods play a pivotal role in addressing data scarcity and imbalance, thereby improving model performance and generalization.
As shown in Fig. \ref{summary}, we summarize the development and evolution of data augmentation and synthesis techniques in recent years. 
This section primarily introduces the current classification of data generation methods, distinguishing between \textbf{data augmentation}, which enhances existing data samples through transformations, and \textbf{data synthesis}, which creates entirely new samples from scratch or based on generative models. Both methods differ in their approach to acquiring data but aim to expand datasets. Furthermore, data augmentation and synthesis methods can be categorized into subclasses from multiple dimensions.
Each approach has unique strengths and applications, enabling researchers to tailor their data generation strategies to specific needs and goals.

\subsection{Data Augmentation}
\label{subsec:data augmentation}
Data augmentation, a type of generation approach from data to data, generally involves manipulating the original data to increase its diversity and quantity without significantly altering its essential characteristics. Techniques used in data augmentation are designed to enhance the richness of existing data samples through transformations or perturbations. Across different modalities, data augmentation techniques often exhibit similarities. For instance, in image data, augmentation operations encompass mosaic\cite{Jocher_YOLOv5_by_Ultralytics_2020}, flipping\cite{shorten2019survey}, copy-pasting\cite{ghiasi2021simple}, adding noise\cite{maharana2022review}, pairing\cite{inoue2018data} and so forth. Similarly, in text data, augmentation operations involve synonym replacement\cite{kobayashi2018contextual}, copy-pasting\cite{shorten2021text}, etc. Moreover, to cater to the demands of multimodal learning, existing research has addressed cross-modal information alignment during data augmentation. MixGen\cite{hao2023mixgen} generates new training samples by linearly interpolating images and concatenating text sequences from two existing image-text pairs. The semantic relationship within the newly generated image-text pair remains consistent and matched. Recently, in the rapidly advancing landscape of LLMs, data augmentation has emerged as a cornerstone for bolstering model performance through the diversification of training exemplars, circumventing the necessity for extensive additional data gathering. From a data-centric perspective, we systematically categorize existing research on data augmentation into three distinct categories: \textbf{data labeling}\cite{liu2022mind,tornberg2023chatgpt,zhu2023can,gilardi2023chatgpt,alizadeh2023open,khan2023q}, \textbf{data reformation}\cite{yoo2021gpt3mix,dixit2022core,dunlap2023diversify,lu2024llamax}, and \textbf{co-annotation}\cite{li2023coannotating,bertaglia2023closing,ding2024data}.

\subsubsection{\textbf{Data Labeling}}
\label{subsubsec:data labeling}
Data labeling endeavors to leverage the comprehensive language understanding capabilities of LLMs to annotate vast unlabeled datasets. This methodology is particularly beneficial in fields that possess a substantial unlabeled data corpus, encompassing domains such as cross-lingual processing and multimodal learning\cite{zhu2023can,gilardi2023chatgpt,alizadeh2023open}, where the automation of annotation can significantly expedite the data preparation process. Recent research studies the zero-shot annotation ability of LLMs, such as GPT-4 on labeling political Twitter\cite{tornberg2023chatgpt}. Moreover, Khan et al. \cite{khan2023q} focus on Visual question answering (VQA) tasks by generating pseudo-label data from unlabeled images by utilizing the SelTDA framework.

\subsubsection{\textbf{Data Reformation}}
\label{subsubsec:data reformation}
Data reformation involves transforming and restructuring existing data into a broader spectrum of variations, thereby facilitating more fine-grained data augmentation\cite{dixit2022core,dunlap2023diversify}. This approach aims to enrich the training landscape with diverse yet pertinent examples, enhancing the model's robustness and generalization capabilities. Classic methods such as rotation\cite{kalra2021towards}, color channel transformation\cite{gillespie1987color}, and synonym replacement\cite{kobayashi2018contextual} are commonly used. Recently, approaches utilizing LLMs have also emerged. For example, Chen et al.\cite{chen2022disco} propose Disco, an approach that harnesses LLMs to produce large-scale, high-quality counterfactual data.

\subsubsection{\textbf{Co-Annotation}}
\label{subsubsec:co-annotation}
Co-annotation designates the collaborative effort between human annotators and LLMs in the annotation process\cite{bertaglia2023closing}. By integrating the strengths of both annotation methodologies, co-annotation not only mitigates annotation costs but also concurrently enhances annotation performance, fostering a more efficient and effective approach to data annotation. Li et al.\cite{li2023coannotating} introduce CoAnnotating, a framework that strategically assigns data points for annotation either to humans or to LLMs, based on an assessment of the LLM's annotation uncertainty. 

\subsection{Data Synthesis}
\label{subsec:data synthesis}
Data synthesis, on the other hand, aims to create entirely new data from scratch or based on generative models, which are similar to the distribution of real data. 
In recent years, with the explosion and advancements in generative AI\cite{ho2020denoising,dhariwal2021diffusion,rezende2015variational,brown2020language,devlin2018bert,liu2019roberta,pinaya2023generative}, there have been significant strides in the quality and generation efficiency of synthetic data. Based on the requirements of LMs, this paper categorizes data synthesis methods into three main types: \textbf{general model distillation}\cite{eldan2023tinystories,li2023textbooks,chen2023alpagasus,zhao2023genixer,zhang2024multimodal}, \textbf{domain model distillation}\cite{luo2023wizardmath,wei2024magicoder,luo2024wizardcoder,lewkowycz2022solving}, and \textbf{model self-improvement}\cite{maini2024rephrasing,wang2023self,fang2024vila,zelikman2022star}.

\subsubsection{\textbf{General Model Distillation}}
\label{subsubsec:general model distillation}
Among these, general model distillation involves leveraging powerful general models, typically featuring larger parameters and superior performance, such as StableVicuna, ChatGPT, and GPT-4, to generate datasets that can enhance the capabilities of weaker models. There are various ways to employ these powerful models, such as using predefined templates to generate tiny stories\cite{eldan2023tinystories} and leveraging the LLMs themselves to evaluate the quality of the generated data. Phi-1 and its series \cite{gunasekar2023textbooks,li2023textbooks}have demonstrated that a small amount of high-quality data can also train a powerful model, by leveraging the comprehensive generation of textbooks and exercises from GPT-3.5. Some other methods have also achieved performance improvements by generating instruction datasets and fine-tuning models after improving the quality of these datasets\cite{honovich2022unnatural,Taori2023alpaca,chen2023alpagasus}.

\subsubsection{\textbf{Domain Model Distillation}}
\label{subsubsec:domain-specific model distillation}
Domain model distillation pertains to the utilization of models that are tailored to generate data within a particular domain. This approach is often necessary when general models fail to meet the specific needs of industry applications. For instance, in the context of code programming, domain model distillation can be employed to generate instructional data tailored to specific coding tasks\cite{wei2024magicoder,luo2024wizardcoder}. In the realm of mathematics, methods such as Minerva\cite{lewkowycz2022solving} and DeepSeekMath\cite{xin2024deepseekprover} are designed to generate solutions to mathematical problems while ensuring their accuracy and diversity. Additionally, the realm of industry data often presents barriers, such as limited data scales and the inaccessibility of data within specific enterprises within the domain. These factors necessitate the adoption of domain-specific models that can effectively address the unique challenges posed by these scenarios.

\subsubsection{\textbf{Model Self-Improvement}}
\label{subsubsec:model self-improvement}
Model self-improvement refers to the process where a model generates higher-quality data to enhance its capabilities. For instance, leveraging existing instructions to adjust the model and prompting it to paraphrase documents on the web in specific styles, such as Wikipedia-style or QA-style, can be used to jointly pre-train LLMs for both authentic and synthetic paraphrasing tasks\cite{maini2024rephrasing}. Self-Instruct \cite{wang2023self}enhances LMs themselves by autogenerating and refining instructional data, boosting performance with minimal human intervention.

\section{Data Synthesis and Augmentation in the Full Lifecycle of LLM}
\label{sec:full lifecycle}

From the perspective of the full lifecycle of LLM, We divide the existing investigations into six stages, including data preparation, pre-training, fine-tuning, instruction-tuning, preference alignment, and applications. The present section introduces relevant research in each stage.

\subsection{Data Preparation}
\label{subsec:data preparation}
In the data preparation phase, data synthesis and augmentation aim to generate diverse and high-quality datasets for the training of LLMs, addressing the challenge of the scarcity of real-world data.
According to the taxonomy discussed in Section 2, We divide the present subsection into general model distillation and data augmentation.

\begin{table*}
\caption{Data synthesis and augmentation in data preparation. In the table, method outlines the techniques presented by each research. Data source and synthetic data indicate the original data used to generate synthetic data and the synthetic data created for training purposes, respectively. A dash (-) in any cell denotes that the respective content was not mentioned in the cited literature.}
\centering
{\scriptsize
\begin{tabular}{cccccc}
\toprule
 \textbf{} & \textbf{Category} & \textbf{Method} & \textbf{Data Source} & \textbf{Synthetic Data} & \textbf{Date} \\
\hline
\multirow{19}{*}{General model distillation} & \multirow{13}{*}{Uni-modality} & - & 1500 basic words & TinyStories\cite{eldan2023tinystories} &  05/2023\\
    ~ & ~ & Genie\cite{yehudai2024genie} & ELI5, NQ and ASQA & Wish-QA, etc. & 01/2024\\
    ~ & ~ & Self-Instruct\cite{wang2023self} & 175 human-written samples & 52k instructions & 12/2022\\
    ~ & ~ & Evol-Instruct\cite{xu2023wizardlm} & Alpaca & Evol-instruct data & 04/2023\\
    ~ & ~ & Iterative Question Composing\cite{liu2024augmenting} & MATH, MeTaMathQA & MMIQC & 01/2024 \\
    ~ & ~ & - & MATH, AQuA, etc. & MathInstruct\cite{yue2023mammoth} & 09/2023\\
    ~ & ~ & - & Stack & Case2Code\cite{shao2024case2code} & 07/2024 \\ 
    ~ & ~ & OSS-Instruct\cite{wei2024magicoder} & starcoderdata & OSS-Instruct data & 12/2023\\
    ~ & ~ & DIALOGIC\cite{li2022controllable} & MultiWOZ & task-oriented dialogues & 10/2022 \\
    ~ & ~ & - & Meta topics, materials, etc. & UltraChat\cite{ding2023enhancing} & 05/2023 \\
    ~ & ~ & MAGPIE\cite{xu2024magpie} & predefined instruction template & MAGPIE-Air, etc. & 06/2024\\
    ~ & ~ & Generator prompts\cite{chen2024genqa} & hand-written meta-prompts & GenQA & 06/2024\\ 
    \cline{2-6}
    ~ & \multirow{6}{*}{multi-modality}& Genixer\cite{zhao2023genixer} & VQAv2, VG, etc. & Genixer-915K & 12/2023\\
    ~ & ~ & Multimodal self-instruct\cite{zhang2024multimodal} & key word seeds & multimodal self-instruct dataset & 07/2024\\
    ~ & ~ & - & specific characteristics & ChartLlama\cite{han2023chartllama} & 11/2023 \\ 
    ~ & ~ & - & Flickr30k, Visual Genome & ComVint\cite{du2023makes} & 11/2023\\
    ~ & ~ & AnyGPT\cite{zhan2024anygpt} & 100 meta topics & AnyInstruct-108k & 02/2024 \\
    ~ & ~ & - & 100k images & ShareGPT-4V\cite{chen2023sharegpt4v} & 11/2023 \\
    \hline
\multirow{14}{*}{Data augmentation} & \multirow{13}{*}{Uni-modality} & FullAnno\cite{Hao2024FullAnno} & Coco, Visual Genome & re-annotated COCO and Visual Genome & 09/2024\\
    ~ & ~ & - & MultiWOZ, etc. & MultiWOZ one-shot, etc.\cite{labruna2023unraveling} & 05/2023\\
    ~ & ~ & - & GSM8k & TinyGSM\cite{liu2023tinygsm} & 12/2023\\
    ~ & ~ & GPT3Mix\cite{yoo2021gpt3mix} & SST-2, CR, et. & RT20 & 04/2021 \\
    ~ & ~ & CORE\cite{dixit2022core} & IMDB & augmented IMDB & 10/2022\\
    ~ & ~ & DISCO\cite{chen2022disco} & SNLI & DISCO dataset & 12/2022 \\
    ~ & ~ & Co-Annotating\cite{li2023coannotating} & six classification datasets & co-annotated dataset & 10/2023\\
    ~ & ~ & Dialgen\cite{lu2023dialgen} & AIC & Dialgen-AIC & 07/2023\\
    ~ & ~ & ToolCoder\cite{zhang2023toolcoder} & CodeSearchNet-Pyhthon & 53k augmented data & 05/2023\\
    ~ & ~ & - & 100 hand-designed scripts & AMPS\cite{hendrycks2021measuring} & 03/2021\\
    ~ & ~ & Mind's Eye\cite{liu2022mind} & physical reasoning questions & UTOPIA & 10/2022\\
    ~ & ~ & - & scientific papers, web data & Proof-Pile-2\cite{azerbayev2023llemma} & 10/2023\\
    ~ & ~ & data pruning\cite{tsai2024code} & MBPP, HumanEval & pruning dataset & 07/2024\\
    \cline{2-6}
    ~ & \multirow{1}{*}{multi-modality}& ALIA\cite{dunlap2023diversify} & CUB & task-relevant synthetic images & 05/2023 \\
    \bottomrule
\end{tabular}}
\label{tab:data preparation}
\end{table*}

\subsubsection{\textbf{General Model Distillation.}}
This way aims to leverage the powerful capabilities of general LLMs to distill high-quality data.
According to the approach and data modality, we further divided general model distillation into five categories: synthesize from seeds, synthesize reasoning steps, synthesize with controllability, synthesize from scratch, and synthesize multimodal data.
 
\textbf{Synthesize from Seeds.}
To synthesize datasets for specific tasks, prompting LLMs with a small number of relevant examples can effectively produce high-quality datasets at a low cost.
For instance, to investigate “how small can an LLM be to achieve certain capabilities”, TinyStories\cite{eldan2023tinystories} is constructed by instructing an LLM to generate stories that combine three words randomly chosen from 1500 basic words, and it can be used to train and evaluate language models.
Based on the collected large-scale functions, Case2Code\cite{shao2024case2code} incorporates LLMs to generate suitable inputs for these functions and utilizes the code interpreter to calculate their corresponding outputs.
Due to the potential insufficiency in quantity and diversity of single-round synthetic data, methods for iterative data synthesis are investigated.
For example, Self-Instruct\cite{wang2023self} can be repeated for many iterations to accumulate a substantial volume of tasks. In each iteration, an LLM is prompted to generate new instructions from a small seed set, then creates input-output instances for each instruction independently.
Similarly, Evol-Instruct\cite{xu2023wizardlm} can be conducted over multiple rounds to gather a sufficient dataset encompassing various complexities. In each evolution, in-depth and in-breadth evolving are employed to either enhance the basic instructions to more sophisticated ones or innovate entirely new directives.

\textbf{Synthesize Reasoning Steps.}
To enhance the reasoning capability of LLMs, additional reasoning steps are generated in the process of data synthesis.
The synthetic question-response pairs in MMIQC\cite{liu2024augmenting} are iteratively constructed by augmenting the initial problems and adding additional reasoning steps without altering their intrinsic logical structure.
Similarly, an effective generation strategy is put forward in which an LLM is requested to synthesize chain-of-thought (CoT) answers after question generation and verification\cite{li2024common}.
Based on the generation of question-CoT pairs through Self-Instruct, MathInstruct\cite{yue2023mammoth} further supplements the Program-of-Thought (PoT) rationale to simplify the math-solving process.

\textbf{Synthesize with Controllability.}
To control the quality of synthetic data, researches are conducted into techniques for data synthesis with controllability.
Driven by the goal of reducing the potential bias of synthetic data, 
OSS-Instruct\cite{wei2024magicoder} utilizes open-source seed code snippets to prompt an LLM in generating coding problems and corresponding solutions. The seed snippets provide controllability of the generation and encourage the LLM to synthesize a variety of coding problems.
Similarly, Genie\cite{yehudai2024genie} prompts an LLM with four content-example pairs to generate a synthetic example to match the extracted content with no example.
In addition, seeded with a few annotated dialogues, DIALOGIC\cite{li2022controllable} guides GPT-3 to synthesize annotated dialogues in a controllable way, in which an auxiliary generator and slot-value match filter are utilized to mitigate de-generation and over-generation issues, respectively.

\textbf{Synthesize from Scratch.}
Another approach avoids reliance on the seed dataset, synthesizing data from scratch.
For instance, UltraChat\cite{ding2023enhancing} is composed of questions about the world, creation and generation, and assistance on existing materials, among which questions about the world request ChatGPT to generate meta topics, subtopics, and questions about conceptions from scratch.
As aligned LLMs can generate user queries due to the auto-regression nature, Magpie\cite{xu2024magpie} directly constructs instruction data by prompting aligned LLMs with a pre-query template to generate instructions along with their corresponding responses.
Focusing on generating large instruction datasets from a single prompt, generator prompts\cite{chen2024genqa} boost the output diversity by asking an LLM to produce a long list of possible choices and select one of the candidates at random.

\textbf{Synthesize Multimodal Data.}
Similar to unimodal, Prompting powerful LLMs like GPT to synthesize data based on seed sets is also the most common method for multimodal data synthesis.
For instance, ShareGPT4V\cite{chen2023sharegpt4v} consists of 100K high-quality captions that are produced by employing data-specific prompts to guide GPT4-vision in generating comprehensive descriptions for supervised fine-tuning. Additionally, an alternative caption model is fine-tuned on the 100k high-quality captions, to expand the number of captions for pre-training.
Taking into account the possible issue of overly simplistic cross-modal instructions generated by LLMs, ComVint\cite{du2023makes}, when presented with an image that has available annotations, adopts a pipeline that includes synthesis, complication, and reformulation.
Standing distinct from the previous data synthesis approach, StableLlava\cite{li2023stablellava} synchronously synthesizes images and dialogues.
The method first employs ChatGPT to craft prompts for image generation and develops content-rich dialogues, subsequently leverages StabeDiffusion to synthesize images from these prompts.
Multimodal data can also be synthesized from scratch.
To create an image and a corresponding instruction from scratch, Multi-modal Self-Instruct\cite{zhang2024multimodal} first instructs the LLM to conceive an original visual concept. Subsequently, it produces detailed code to visualize the idea. Once the desired image is synthesized, the LLMs self-instruct multiple sets of high-quality question-answer pairs tailored to the visual content.
Likewise, in the field of interpreting chart figures, ChartLlama\cite{han2023chartllama} initially leverages the capabilities of GPT-4 to generate chart data by providing specific attributes like topics, distributions, and trends. Following this, GPT is further employed to generate both chart figure and the associated instruction-answer data.
AnyInstruct-108k\cite{zhan2024anygpt} is constructed by a two-stage approach including the generation of text-based conversations incorporating multimodal elements from meta topics and text-to-multimodality conversion.
Additionally, there are methods that empowers MLLM to synthesize data rather than prompting GPT.
Genixer\cite{zhao2023genixer}, a holistic data generation pipeline, consists of four key steps including instruction data collection, instruction template design, empowering MLLMs, and data generation and filtering.

\subsubsection{\textbf{Data Augmentation.}}
\label{subsubsec:DA in preparation}
Data augmentation aims to further process existing data to obtain a more diverse set of high-quality data. In the present survey, we divide the existing methods of data augmentation into four groups: data labeling, data reformation, co-annotation, and non-LLM-driven.

\textbf{Data Labeling.}
Data labeling aims to harness the language comprehension abilities of LLMs for annotating unlabeled datasets\cite{alizadeh2023open, gilardi2023chatgpt}. 
Based on a balanced sample with 500 tweets from Republican and Democratic politicians, ChatGPT-4 is utilized to annotate the political affiliation\cite{tornberg2023chatgpt}. The results indicate that LLM annotations display higher accuracy and lower bias than human classifiers. 
To apply to distinctive annotation tasks, one approach\cite{zhu2023can} begins by creating a generalized prompt template. This template is then utilized to generate a ChatGPT prompt aims at extracting labels in alignment with the dataset's original annotation methodology.
Similarly, another method involves initially creating dialogues that closely match the content of a reference dialogue, subsequently prompting the LLM to label the generated conversation using the same annotation schema provided in the existing repository\cite{labruna2023unraveling}.
Further, scholars have studied methods to improve the quality of LLMs' labeled data using additional information.
For instance, FullAnno\cite{Hao2024FullAnno} instructs LLMs to acquire comprehensive annotations for images with prompts including the category and position of objects, region description, and text information within the image.
In the field of speech emotion recognition, one approach\cite{latif2023can} first annotates samples based solely on text. Afterwards, it enhances the annotation by integrating audio features and gender information alongside textual data. In addition, a VQ-VAE is employed to produce a 64-dimensional discrete representation of the audio, which is supplied to the LLM.

\textbf{Data Reformation.}
Data reformation attempts to transform existing data into a wider array of variations, and it typically involves the utilization of prompt engineering to guide LLMs in generating reformatted data. 
For example, TinyGSM\cite{liu2023tinygsm} is constructed by prompting an LLM to generate problem variants from GSM8K, subsequently filtering out low-quality instances.
Similarly, GPT3Mix\cite{yoo2021gpt3mix} extracts augmentation from the generation of the LLM by constructing a task-specific prompt from the selected examples and meta-information about the dataset.
Data reformation is also widely applied in generating counterfactual augmented data.
Specifically, CORE\cite{dixit2022core} initially learns to retrieve relevant text excerpts. Following this the retrieved excerpts along with instructions and demonstrations are supplied as a prompt to GPT-3 to counterfactually edit input text.
In addition, DISCO\cite{chen2022disco} firstly decomposes given task instances into spans using linguistic processing tools, and subsequently employs prompt engineering and in-context learning with an LLM to overgenerate a diverse set of perturbations for these instances.
For multi-modality, ALIA\cite{dunlap2023diversify} first generates captions for each image, and summarizes the captions into a short list of domain descriptions with an LLM. Then uses these descriptions to generate edits of the training data with Stable Diffusion.

\textbf{Co-Annotation.}
Co-annotation refers to the process where humans and LLMs annotate unlabeled data together.
For instance, Toolcoder\cite{zhang2023toolcoder} utilizes human-written input-output within the prompt to direct ChatGPT in annotating tool-augmented dataset.
To address the problem of low agreement between annotators, ChatGPT is used to augment the annotation process with phrases identified as relevant features and brief explanations\cite{bertaglia2023closing}.
Investigations on allocating data within the same dataset to humans and ChatGPT can achieve a more efficient and more accurate annotation.
Specifically, the CoAnnotating\cite{li2023coannotating} framework automatically decides whether each data instance should be annotated by humans or by the LLMs by computing the uncertainty level of the LLMs’ annotations. 
There are also iterative co-annotation methods for data augmentation.
For example, initiated with a task prompt consisting of a description for desired dialogue and dialogue history, Dialgen\cite{lu2023dialgen} first proposes a candidate subdialogue, subsequently validates, edits and annotates teh generated subdialogue by human reviewers before requesting continuation via an updated prompt to the LLM. The process can be repeated multiple times until the dialogue is complete.

\textbf{Non LLM-Driven.}
Some methods do not use LLMs to synthesize or filter high-quality data.
For exmaple, AMPS\cite{hendrycks2021measuring}, an extensive and varied corpus for mathematics pretraining, comprises over 5 million problems generated with Mathematica scripts, based on 100 meticulously crafted modules. 
In the field of physics, Mind’s Eye\cite{liu2022mind} utilizes a computational physics engine to produce ground-truth answers for the UTOPIA multi-task physics alignment dataset, designed to assess the ability of LLMs to understand fundamental physical laws. 
Besides, filtering and pruning strategy are utilized for data augmentation.
For instance, Proof-Pile-2\cite{azerbayev2023llemma} is an extensive dataset with 55B tokens of mathematics and mathematical code, enriched by filtering high-quality data from publicly available resources.
Recognizing substantial redundancies in synthetic training data, an efficient and scalable pruning strategy\cite{tsai2024code} is proposed which encompasses encoding, dimensionality reduction, clustering, and pruning.

\subsection{Pre-Training}
\label{subsec:pre-training}
During the stage of pre-training, data synthesis and augmentation can provide LLMs with abundant, diverse, and controllable training data cost-effectively and efficiently, thereby enhancing model performance and reducing bias.
We also discuss the existing methods from three perspective: model self-improvement, general model distillation, and data augmentation.

\begin{table*}
\caption{Data synthesis and augmentation in pre-training. Method outlines the techniques presented by each research. Data source and synthetic data indicate the original data used to generate synthetic data and the synthetic data created for pre-training, respectively. Base model and pre-trained model indicate the foundational models and the models that have undergone pre-training, respectively. A dash (-) in any cell denotes that the respective content was not mentioned in the cited literature.}
\centering
{\scriptsize
\begin{tabular}{cccccccc}
\toprule
 \textbf{} & \textbf{Category} & \textbf{Method} & \textbf{Data Source} & \textbf{Synthetic Data} & \textbf{Base Model} & \textbf{Pre-trained Model} & \textbf{Date} \\
\hline

\multirow{1}{*}{Model self-improvement} & \multirow{1}{*}{Multi-modality} & - & MMC4, Coyo, ShareGPT4v & Images with re-captioned texts & VILA & VILA-2\cite{fang2024vila} & 07/2024 \\
\hline
\multirow{7}{*}{model} & \multirow{4}{*}{Uni-modality} & - & code snippets & 1B tokens Python textbooks & 1.3B parameter model & phi-1-Base\cite{gunasekar2023textbooks} & 06/2023 \\
    ~ & ~ & - & 20k topics & 20B tokens textbooks data  & 1.3B parameter model & phi-1.5\cite{li2023textbooks}  & 09/2023 \\
    General & ~ & - & conversation seed & TinyDialogues\cite{feng2024child} & GPT-2 & - & 08/2024 \\
    ~ & ~ & - & Scientific corpora & SciLitIns & Qwen2-Base & SciLitLLM\cite{li2024scilitllm} & 08/2024 \\
    distillation & ~ & TRAIT\cite{liang2024task} & GSM8k, OpenWebMath & Task-oriented synthetic passages & Mistral & - & 06/2024 \\
    \cline{2-8}
    ~ & \multirow{1}{*}{Multi-modality} & - & 100 meta topics  & AnyInstruct-108k & Llama 2 & AnyGPT\cite{zhan2024anygpt} & 02/2024 \\
    \hline
\multirow{4}{*}{Augmentation} & \multirow{4}{*}{Uni-modality} & - & sub-molecule groups & Physics-based synthetic data\cite{liu2024large} & MoLFormer & - & 07/2024 \\
    ~ & ~ & WRAP\cite{maini2024rephrasing} & C4 & Re-phrased texts & decoder-only transformers & - & 01/2024 \\
    ~ & ~ & - & CC100 & knowledge-intensive multilingual data & mBERT, XLM-R & KMLM\cite{liu2021enhancing} & 11/2021 \\
    ~ & ~ & - & bioS & bioR\cite{zhu2023physics} & Llama & - & 09/2023 \\
    \bottomrule
\end{tabular}}
\label{tab:pre-training}
\end{table*}

\subsubsection{\textbf{Model Self-Improvement.}}
In the pre-training phase, model self-improvement denotes synthesize data by an LLM, and further utilizes the synthetic data to pre-train the same LLM. For instance, VILA-2\cite{fang2024vila} utilize a self-augmenting process where the present round of VILA is used to generate long and detailed captions with appropriate prompt choice and conversation template for the next round pretraining.

\subsubsection{\textbf{General Model Distillation.}}
General model distillation denotes the utilization of a general LLM with a strong capability to distill high-quality data.
To demonstrate the power of high-quality data in breaking existing scaling laws, phi-1\cite{gunasekar2023textbooks} is pre-trained on code dataset of “textbook quality”, both synthetically generated with GPT-3.5 and filtered from web sources. Following the method of phi-1 and extending the task to common sense reasoning in natural language, phi-1.5\cite{li2023textbooks} is pre-trained on a combination of phi-1’s training data and newly created synthetic data.
Inspired by TinyStories, TinyDialogues\cite{feng2024child} is created by prompting GPT-4 to generate realistic dialogues featuring children of various ages as the main participants.
In the continual pre-training stage of SciLitLLM\cite{li2024scilitllm}, Llama3-7B-Instruct is utilized to correct the errors introduced during the PDF parsing process followed by supervised transfer learning on a classifier to filter out low-quality texts from the dataset.
For multi-modality, AnyGPT\cite{zhan2024anygpt}, pre-trained on a multimodal text-centric dataset, is an any-to-any multimodal language model capable of comprehending and producing diverse modalities. Utilizing the open-source text-to-image generation model GLIDE, it has been demonstrated that synthetic data significantly aids classifier learning and holds substantial promise for model pre-training\cite{he2022synthetic}.

\subsubsection{\textbf{Data Augmentation.}}
\label{subsubsec:DA in pre-training}
Data augmentation aims to further process existing data to obtain a more diverse dataset. In the pre-training stage, there are mainly two kinds of methods: data reformation and non-LLMs-driven method.


\textbf{Data Reformation.}
Data reformation transforms the original dataset to obtain a new dataset with diversity and quality.
For instance, WRAP\cite{maini2024rephrasing} employs an off-the-shelf instruction-tuned model to paraphrase web documents in various styles, thereby pre-training LLMs on a combination of real and synthetic rephrases.
Similarly, the BIO dataset bioR\cite{zhu2023physics} is constructed by rewriting the synthetic dataset of 100k  biographies using Llama to make them close to real-life biography style.


\textbf{Non LLMs-Driven.}
Other methods augment the original dataset without the utilization of LLMs.
For example, Code Llama undergoes further pretraining on Proof-Pile-2, a 55B-token augmented dataset consisting of scientific papers and web data, yielding LLEMMA\cite{azerbayev2023llemma}. 
KMLM\cite{liu2021enhancing} is pre-trained on massive multilingual knowledge graph triples which are constructed by converting the structured knowledge from knowledge graphs to sequential data. 
To tackle the pathology of data scarcity, a physics-based modeling framework\cite{liu2024large} is proposed that generates a multitude of synthetic data to align the LLM to a physically consistent initial state.

\subsection{Fine-Tuning}
\label{subsec:fine-tuning}
In the fine-tuning phase, data synthesis and augmentation refer to the employment of the generated data to fine-tune LLMs. It has been proven that generated data can effectively contribute to the fine-tuning of LLMs\cite{wang2023self, xu2023wizardlm}. 
We discuss the existing methods from three perspectives: model self-improvement, general model distillation, and data augmentation.

\begin{table*}
\caption{Data synthesis and augmentation in fine-tuning. In the table, method outlines the techniques presented by each research. Data source and synthetic data indicate the original data used to generate synthetic data and the synthetic data created for fine-tuning, respectively. Base model and fine-tuned model indicate the foundational models and the models that have undergone fine-tuning, respectively. A dash (-) in any cell denotes that the respective content was not mentioned in the cited literature.}
\centering
{\scriptsize
\begin{tabular}{cccccccc}
\toprule
\textbf{} & \textbf{Category} & \textbf{Method} & \textbf{Data Source} & \textbf{Synthetic Data} & \textbf{Base Model} & \textbf{Fine-tuned Model} & \textbf{Date} \\
\hline
\multirow{11}{*}{self-} & \multirow{11}{*}{Uni-modality} & STaR\cite{zelikman2022star} & Arithmetic, etc. & A rationale dataset & GPT-J & - & 03/2022 \\
    ~ & ~ & - & GSM8k, etc. & reasoning dataset & PaLM-540B & LMSI\cite{huang2022large} & 10/2022 \\
    ~ & ~ & ReST\cite{gulcehre2023reinforced} & IWSLT 2014, etc.& - & Transformer & - & 08/2023 \\
    ~ & ~ & ReST-{EM}\cite{singh2023beyond} & MATH, APPS & synthetic math and code & PaLM 2 & - & 12/2023 \\
    Model & ~ & - & RLHF V5 & Self-instruction dataset & Llama 2 & Code Llama-Instruct\cite{roziereCodeLlamaOpen2024} & 08/2023 \\
    ~ & ~ & - & miniF2F, FIMO & theorem-proof pairs & DeepSeek math & DeepSeek Prover\cite{xin2024deepseekprover} & 05/2024 \\
    improvement & ~ & Self-Translate\cite{ri2024self} & SQuAD, multiNLI & Translated synthetic data & Llama 2 & - & 06/2024 \\
    ~ & ~ & RFT\cite{yuan2023scaling} & GSM8k & Reject sampling samples & Llama & - & 08/2023 \\
    ~ & ~ & CodeRL\cite{le2022coderl} & public code & synthetic samples & CodeT5 & - & 07/2022 \\
    ~ & ~ & SPIN\cite{chen2024self} & Ultrachat200k & 100k synthetic dataset & zephyr & - & 01/2024\\
    ~ & ~ & Self-Instruct\cite{wang2023self} & 175 human-written samples & 52k instructions & GPT-3 & GPT-3-{self-inst} & 12/2022 \\
\hline
\multirow{19}{*}{model} & \multirow{15}{*}{Uni-modality} & Impossible Distillation\cite{jung2024impossible} & contextual constraints & DIMPLE & T5-large & impossible-T5 & 05/2023 \\
    ~ & ~ & - & 175 instruction-response pairs & instruction-following samples & Llama & Alpaca\cite{Taori2023alpaca} & 2023 \\
    ~ & ~ & LAB\cite{sudalairaj2024lab} & a taxonomy & synthetic instruction dataset & Llama-2, Mistral & Labradorite, Merlinite & 03/2024 \\
    ~ & ~ & GLAN\cite{li2024synthetic} & a taxonomy & synthetic instruction dataset & Mistral & - & 02/2024 \\
    ~ & ~ & - & Code Alpaca & instruction-following data & StarCoder & WizardCoder\cite{luo2024wizardcoder} & 06/2023 \\
    ~ & ~ & CLINGEN\cite{xu2024knowledgeinfused} & clinically relevant knowledge & synthetic clinical data & PubMedBERT & - & 11/2023 \\
    ~ & ~ & Self-chat & MedQuAD & 111.5k dialogues & Llama & Baize\cite{xu2023baize} & 04/2023 \\
    ~ & ~ & LLM2LLM\cite{lee2024llm2llm} & GSM8k, etc. & task-specific datasets & Llama 2 & - & 03/2024 \\
    General & ~ & - & CMeKG & Question-answer instances & Llama & HuaTuo\cite{wang2023huatuo} & 04/2023 \\
    ~ & ~ & - & FLAN-v2 & Query-response pairs & Llama & Orca\cite{mukherjee2023orca} & 06/2023 \\
    distillation & ~ & - & FLAN-v2, Orca 2 dataset & 1 million data & Llama 2 & Orca 2\cite{mitra2023orca} & 11/2023 \\
    ~ & ~ & Evol-Instruct\cite{xu2023wizardlm} & Alpaca & 250k instructions & Llama & WizardLM & 04/2023 \\
    ~ & ~ & OSS-Instruct\cite{wei2024magicoder} & starcoderdata & coding instruction data & CodeLlama, etc. & Magicoder & 12/2023 \\
    ~ & ~ & - & MATH, etc. & MathInstruct & Llama & MAmmoTH\cite{yue2023mammoth} & 09/2023 \\
    \cline{2-8}
    ~ & \multirow{5}{*}{Multi-modality} & NExT-GPT\cite{wu2023next} & multi-modal data & MosIT dataset & - & - & 09/2023\\
    ~ & ~ & - & COCO & instruction-following data & VIcuna & LLava\cite{liu2023llava} & 04/2023\\
    ~ & ~ & - & PMC-15M & instruction-following data & LLaVA & LLaVA-Med\cite{li2024llava} & 06/2023 \\
    ~ & ~ & - & specific characteristics & ChartLlama & LLaVA & ChartLlama\cite{han2023chartllama} & 11/2023 \\
    ~ & ~ & - & 100k images & ShareGPT-4V & LLaVA & ShareGPT4V\cite{chen2023sharegpt4v} & 11/2023 \\
    \hline
\multirow{9}{*}{Augmentation} & \multirow{8}{*}{Uni-modality} & - & NCBI Disease, GAD & structured information\cite{tang2023does} & BERT, etc. & - & 03/2023\\
    ~ & ~ & - & MedQA & UltraMedical & Llama 3 & Llama-3-UltraMedical\cite{zhang2024ultramedical} & 06/2024 \\
    ~ & ~ & - & GSM8k, MATH & MetaMathQA & Llama & MetaMath\cite{yu2024metamath} & 09/2023 \\
    ~ & ~ & Symbol tuning\cite{wei2023symbol} & NLP datasets & input-label pairs & Flan-PaLM & - & 05/2023 \\
    ~ & ~ & - & MedMCQA & DISC-Med-SFT & Baichuan-13B-Base & DISC-MedLLM\cite{bao2023discmedllm} & 08/2023 \\
    ~ & ~ & MathGenie\cite{lu2024mathgenie} & GSM9k, MATH & problem-solution pairs & Llama 2 & - & 02/2024 \\
    ~ & ~ & - & scientific papers, web data & Proof-Pile-2 & Code Llama & Llemma\cite{azerbayev2023llemma} & 10/2023 \\
    ~ & ~ & - & MedDialog-CN, IMCS-V2, etc. & BianQueCorpus & ChatGLM & BianQue\cite{chen2023bianque} & 10/2023 \\
    \cline{2-8}
    ~ & \multirow{1}{*}{Multi-modality} & SelTDA\cite{khan2023q} & A-OKVQA, AQUA & question-answer pairs & ViT-B/16 & - & 2023\\
    \bottomrule
\end{tabular}}
\label{tab:fine-tuning}
\end{table*}

\subsubsection{\textbf{Model Self-Improvement.}}
The model self-improvement approach enables the LLM to learn from its outputs through a feedback process, thus eliminating the need for external support. Based on whether the method uses iterative self-improvement and the modality of synthetic data, we group the existing self-improvement strategy into three categories: single-shot self-improvement, iterative self-improvement, and multi-modal self-improvement.

\textbf{Single-Shot Self-Improvement.}
Single-shot self-improvement denotes the process of synthesizing data through an LLM and then performing a single fine-tuning for the same LLM with the synthesized data.
One category of methods involves supplementing information to the training dataset.
For example, one approach\cite{huang2022large} involves using an LLM to create ‘high-confidence’ rationale-augmented responses for unlabeled questions through Chain-of-Thought prompting and self-consistency checks. 
Based on the observation that the model performance has a log-linear relation versus the supervised data, reject sampling fine-tuning (RFT)\cite{yuan2023scaling} utilizes the LLM's capabilities to generate and compile accurate reasoning trajectories as an augmented dataset. 
Similarly, Self-Translate-Train\cite{ri2024self} capitalizes on the LLM’s translation prowess to produce synthetic training datasets in the desired target language, subsequently employing this self-generated data for fine-tuning. 
On the other hand, another category of methods synthesizes new samples based on existing seed data.
For instance, CodeRL\cite{le2022coderl} introduces an actor-critic framework that employs an LLM as the actor network to produce synthetic samples, while a separate neural network serves as a critic model to assess the quality of these samples. 
Further, Self-Instruct\cite{wang2023self} prompts an LLM to generate new instructions and corresponding instances which can be used for the instruction tuning of the LLM itself.
Based on the self-instruct, Code Llama-Instruct\cite{roziereCodeLlamaOpen2024} are enhanced through fine-tuning on a combination of proprietary instructional data and synthetic self-instruct dataset which is crafted by prompting Llama 2 for coding issues and Code Llama for corresponding unit tests and solutions. 

\textbf{Iterative Self-Improvement.}
Furthermore, to improve the quality, diversity and amount of synthetic data, various approaches iteratively synthesize datasets and continuously fine-tune LLM in improved loops. 
To this end, a self-improvement strategy\cite{haluptzok2022language} is proposed where the LLM generates its own synthetic puzzle-solution pairs which are filtered before being used to fine-tune the LLM itself. 
Further, STaR\cite{zelikman2022star} constructs an augmented dataset by using the LLM’s rationale generation ability and justifying ground-truth answers to problems the model failed to solve. 
Besides, ReST\cite{gulcehre2023reinforced} first augments the training dataset by generating multiple output predictions using an LLM, and then fine-tunes the same LLM on the filtered dataset with an offline reinforcement learning objective.
Following ReST, ReST-EM\cite{singh2023beyond} refrains from augmenting the dataset in the generate step with human-generated outputs and fine-tunes the base LLM instead of the model obtained from the previous iteration in the improve step. 
In addition, DeepSeek-Prover\cite{xin2024deepseek} is consistently fine-tuned on the synthetic data which is generated through auto formalization, quality filtering, and statement proving, and the updated model is then utilized for the subsequent iteration. 
In SPIN\cite{chen2024self}, a self-play strategy is implemented where an LLM is fine-tuned to distinguish the response of the opponent player (the LLM from the previous iteration) from the target data distribution, thereby iteratively aligning the LLM with the target data distribution.

\subsubsection{\textbf{General Model Distillation.}}
General model distillation denotes distilling high-quality fine-tuning data from a powerful LLM.
In the present survey, we divide the existing method of general model distillation into five categories: synthesize with seeds, synthesize data iteratively, synthesize reasoning steps, taxonomy-driven synthesize, and synthesize multimodal data.

\textbf{Synthesize with Seeds.}
Synthesizing data from existing instance examples or data seed is the most common way\cite{Taori2023alpaca, chaudhary2023code, wei2024magicoder}.
For instance, Unnatural Instruction\cite{honovich2022unnatural} gathers examples by initially providing a language model with three seed examples to generate a fourth. The dataset is then expanded by prompting the model to rephrase each instruction. 
By utilizing self-chat to generate a multi-turn chat corpus with ChatGPT, Baize\cite{xu2023baize} is obtained through parameter-efficient tuning and self-distillation with feedback. 
Moreover, CLINGEN\cite{xu2024knowledgeinfused} capitalizes on clinical knowledge extraction to contextualize prompts. The strategy encompasses creating clinical topics from both knowledge graphs and LLMs, as well as extracting writing style recommendations from LLMs. 
Similarly, HuaTuo\cite{wang2023huatuo} is a Llama-based model that has undergone SFT using generated QA which is synthesized through extracting knowledge instances from a knowledge graph and creating additional instances with the help of ChatGPT. 
In addition, Impossible Distillation\cite{jung2024impossible} enhances small, low-quality language models by utilizing paraphrase proximity and critic-guided distillation to create a high-quality paraphrase dataset.

\textbf{Synthesize Data Iteratively.}
To construct high-quality data with diversity, some approaches build frameworks that can be performed multiple times.
For instance, WizardLM\cite{xu2023wizardlm}, fine-tuned with 70k synthetic data generated by the Evol-Instruct method, achieves state-of-the-art results on high-complexity tasks and remains competitive on other metrics. 
Evol-Instruct is further adapted to the code domain with modifications including refining evolutionary instruction, simplifying evolutionary prompts formats, incorporating code debugging and time-space complexity constraints\cite{luo2024wizardcoder}. 
Moreover, LLM2LLM\cite{lee2024llm2llm} fine-tunes a student model on an initial dataset, then identifies errors, and augments training data with synthetic examples from a teacher LLM based on those errors. The process repeats to train the student model on increasingly targeted data points.

\textbf{Synthesize Reasoning Steps.}
Recent studies have concentrated on improving the performance of LLM by imitation learning, drawing with the outputs generated by large foundation models (LFM). However, the smaller language model tends to imitate the style, but not the reasoning process of LFM. To this end, GAIR-Abel\cite{Chern2023generativeAI} found that the structure of the augmented responses significantly impacts the overall performance, with answers that initiate with a paraphrasing of the question and proceed with step-by-step solution showing superior performance than those in standard format. Further, Orca\cite{mukherjee2023orca} is explanation-tuned on augmented query-response pairs with detailed responses from GPT-4. The approach allows Orca to learn from the rich signal including explanation traces, step-by-step thought processes, and other complex instructions. In Orca 2\cite{mitra2023orca}, various reasoning techniques including step-by-step, recall then generate, recall-reason-generate and direct answer are investigated, which endows the model with better reasoning capability.
Additionally, MAmmoTH\cite{yue2023mammoth}, a series of open-source LLMs specifically tailored for general math problem-solving, are fine-tuned on the hybrid instruction tuning dataset MathInstruct with CoT and PoT rationales. 

\textbf{Taxonomy-Driven Synthesize.}
The aforementioned methods are mostly based on synthesizing datasets from seeds, while recent research has adopted another novel approach to synthesizing datasets through a taxonomy-driven method.
To address the issue of tail phenomenon, LAB\cite{sudalairaj2024lab} replaced the random sampling in existing synthetic data generation methods with a taxonomy-driven approach to guide the sampling of synthetic data. Similar to LAB, GLAN\cite{li2024synthetic} utilizes a semi-automatic approach to synthesize large-scale datasets which uses a human-curated taxonomy to generate instruction-tuning data from a teacher model.
Further, SciLitIns\cite{li2024scilitllm} is constructed by a three-step pipeline including collecting a probability table of domain keywords, compiling a list of task descriptions, and prompting GPT-4o to generate.

\textbf{Synthesize MultiModal Data.}
General model distillation also holds great potential for multimodal applications.
For instance, Visual instruction-tuning\cite{liu2024visual} extends instruction-tuning to the language-image multimodal space by leveraging language-only GPT-4 for multimodal instruction-following data generation. 
By fine-tuning LLaVA on multimodal synthetic data, ChartLlama\cite{han2023chartllama} is developed with a broad spectrum of capabilities for chart understanding and generation capabilities.
Besides, A curriculum learning method\cite{li2024llava} is introduced which involves first fine-tuning LLaVA to align biomedical vocabulary and continue training the model using generated instruction-following data by GPT-4. 
Further, ShareGPT4V-7B\cite{chen2023sharegpt4v} is fine-tuned on the ShareGPT4V dataset and demonstrates impressive performance across various multi-modal benchmarks. 
In addition, NExT-GPT\cite{wu2023next} leverages a modality-switching instruction tuning (MosIT) methodology which prompts GPT-4 to generate multimodal dialogues under various scenarios based on template dialogue examples.

\subsubsection{\textbf{Data Augmentation.}}
\label{subsubsec:DA in fine-tuning}
Data augmentation involves enhancing the existing data through various techniques to create a more extensive and varied dataset. In the fine-tuning stage, there are mainly two kinds of methods: data labeling nad data reformation.

\textbf{Data Labeling.}
Data labeling denotes generating annotations for unlabeled data.
For instance, A pipeline\cite{tang2023does} is proposed to generate a large volume of synthetic data with labels using LLMs and further eliminate low-quality or duplicated samples. The results demonstrate the effectiveness of the method compared to LLM’s zero-shot performance.
Futhermore, Llama-3-UltraMedical\cite{zhang2024ultramedical} is obtained by supervised fine-tuning on the UltraMedical dataset, which includes instructions annotated with completions from various LLMs with preferences annotated by GPT-4. 
For mutil-modality, SelTDA\cite{khan2023q} uses the vision language model and target dataset to build a teacher model that can generate question-answer pseudo labels directly conditioned on an image alone.

\textbf{Data Reformation.}
Data reformation refers to transform the existing data into a more diverse form, thereby augmenting the data.
For example, Symbol tuning\cite{wei2023symbol} fine-tunes language model on input-label pairs presented in-context where natural language labels are remapped to arbitrary symbols. 
Further, DISC-MedLLM\cite{bao2023discmedllm} is fine-tuned on the SFT dataset which is constructed by utilizing medical knowledge-graphs, reconstructing real-world dialogues, and rephrasing human-guided preference. 
Besides, MetaMath\cite{yu2024metamath} is fine-tuned on the MetaMathQA dataset which is constructed by rewriting questions with both forward and backward reasoning paths through LLMs. 
In addition, MathGenie\cite{lu2024mathgenie} consists of three components including iterative solution augmentation, question back-translation, and verification-based solution filtering to create diverse and reliable data.
Moreover, BianQueCorpus\cite{chen2023bianque} is constructed through collecting real-world multi-turn health conversations, constructing a data automatic cleaning process, and using ChatGPT to polish the doctors’ suggestion of multi-turn conversations. 


\subsection{Instruction-Tuning}
\label{sec:Instruction-Tuning}
In the instruction tuning phase, data synthetic aims at exploring synthetic instruction or prompt contents to generate instruction-following high-quality data via LLMs. According to the way of synthetic data, they consist of the following categories: (1) general model distillation, (2) model self-improvement, and (3) data augmentation, as shown in Table~\ref{tab:Instruction-Tuning}. 
\begin{table*}
\caption{Data synthesis and augmentation in Instruction-Tuning. A dash (-) indicates no relevant content}
\centering
\setlength{\tabcolsep}{1.0mm}{
{\scriptsize
\begin{tabular}{ccccccc}
\toprule
\textbf{} & \textbf{Category} & \textbf{Method} & \textbf{Data Source} & \textbf{Synthetic Data} & \textbf{Base Mmodel} & \textbf{Target Model}\\
\hline
\multirow{6}{*}{General model distillation,} & \multirow{5}{*}{Uni-modality} & Alpaca \cite{Taori2023alpaca} & 175 instruction-response pairs &  52k instruction-following samples & GPT-3.5 & LLaMA  \\
    ~ &  & Vicuna \cite{chiang2023vicuna}  & 175 instruction-response pairs &  9k instruction-following samples & ChatGPT & LLaMA   \\
    ~ &  & WizardLM \cite{xu2023wizardlm} & 175 instruction-response pairs & 250k instructions & ChatGPT & LLaMA  \\
    ~ & & Orca \cite{mukherjee2023orca} & FLANv2 & 5 million query-response pairs & GPT-4 & LLaMA  \\
    ~ &  & Orca 2 \cite{mitra2023orca} &  FLAN-v2 \& Orca 2 dataset &  1 million data  & GPT-4 & LLaMA   \\
    \cline{2-7}
    ~ & \multirow{1}{*}{Multi-modality} & LLaVA \cite{liu2024visual} & image-text pairs &  text-only prompt-answer pairs & ChatGPT and GPT-4 & Vicuna and CLIP  \\
\hline
\multirow{6}{*}{Model self-improvement} & \multirow{5}{*}{Uni-modality} & Self-Instruct \cite{wang2023self} & 175 human-written samples &  52k instructions & GTP-3 & GTP-3    \\
    ~ &  & Backtranslation \cite{li2023self} & 13.2k instruction examples & 502k instruction-output pairs & LLAMA & LLAMA  \\
    ~ &  & SPIN \cite{chen2024self} & 50k prompts from Ultrachat200k & 100k synthetic dataset & zephyr-7b-sft-full & zephyr-7b-sft-full   \\
    ~ &  & ReST$^{EM}$ \cite{singh2023beyond} & Hendrycks’ MATH and APPS dataset &  32 or 64 solutions per problem & PaLM 2 & PaLM 2   \\
    ~ &  & CAI \cite{bai2022constitutional} &   16 principles or instructions & 318k harmful and helpful instructions & 52B SL-CAI & 52B SL-CAI \\
    \cline{2-7}
    ~ & \multirow{1}{*}{Multi-modality} & SelTDA \cite{khan2023q} & unlabeled images & image-conditional question-answer pairs & ViT-B/16 & ViT-B/16  \\
    \hline
\multirow{9}{*}{Data augmentation} & \multirow{3}{*}{Data labeling} & \cite{tornberg2023chatgpt} & political Twitter messages & Annotating political Twitter messages & Chatgpt-4 &  - \\
& &  Machine Translation \cite{zhang2023prompting} & monolingual data & back-/forward-translation monolingual data &  GLM-130B & - \\
& & T-SciQ \cite{wang2024t} & original question-answer data & planning-based CoT rationales & GPT-3.5 & - \\
    \cline{2-7}
& \multirow{3}{*}{Data reformation} & CORE \cite{dixit2022core} & task-related unlabeled texts & diverse counterfactual perturbations & GPT-3 & -  \\
& &  ALIA \cite{dunlap2023diversify} & image data & language-guided image editing &  GPT-4 &  -\\
& & ChatAug \cite{dai2023chataug} & text data & semantically similar sentences &ChatGPT  &  -\\
    \cline{2-7}
& \multirow{3}{*}{Co-annotation} & CoAnnotating \cite{li2023coannotating} & unstructured texts & responses via different prompt variations  & ChatGPT &  - \\
& &\cite{bertaglia2023closing}  & sponsored content on social media & generated explanations  & ChatGPT  &  -\\
& &ToolCoder \cite{zhang2023toolcoder}  & source code & API-augmented code & ChatGPT & - \\ 
\bottomrule
\end{tabular}}
}
\label{tab:Instruction-Tuning}
\end{table*}

\subsubsection{General Model Distillation}
To obtain diverse data, a popular method adopts a stronger LLM to synthesize data and perform instruction-tuning for a weaker LLM \cite{honovich2022unnatural}, including uni-modal synthesis and multi-modal synthesis. 

\textbf{Uni-Modality.} Uni-modality synthesizes a specific type of data via teacher LLMs.
Alpaca \cite{Taori2023alpaca} first generates instruction-following demonstrations via GPT-3.5 (text-davinci-003) and then fine-tunes llama-7b to create a replicable instruction-following model.
Next, Alpagasus \cite{chen2023alpagasus} discovers that the instruction-tuning dataset used by Alpaca contains many incorrect or irrelevant low-quality instances. In response, they design a quality filtering strategy that leverages powerful LLMs like ChatGPT to automatically identify and filter out low-quality data. The results demonstrated that a small amount of high-quality data was sufficient to train a model with even stronger performance.
Based on Alpaca, Vicuna \cite{chiang2023vicuna} gathers user-shared conversations from ShareGPT.com to build an open-Source chatbot. 
Given that the production of high-complexity instructions may pose a challenge for humans, WizardLM \cite{xu2023wizardlm} proposes an Evol-Instruct strategy to rewrite and produce more complex instructions. Evol-Instruct uses LLMs to automatically mass-produce various instructions at different levels, including two evolution strategies: in-depth evolution and in-breadth evolution. 

While these above models produce abundant data via LLMs, they often lack the ability of reasoning and comprehension skills displayed by LLMs.
To this end, Orca \cite{mukherjee2023orca} and Orca2 \cite{mitra2023orca} imitate the reasoning process of stronger LLMs via explanation traces to output synthetic samples.
Compared to vanilla instruction tuning, Orca leverages system instructions to augment query-response pairs with detailed reasoning explanations. 
Based on Orca, Orca2 further introduces various reasoning strategies, such as step-by-step and recall-reason-generate, to learn to determine the most effective solution strategy for each task.
Orca2 distills a synthetic dataset by collecting FLAN-v2 Collection \cite{longpre2023flan}, 55K Few-Shot dataset, Math dataset \cite{saxton2019analysing}, and 2000 Doctor-Patient Conversations, creating cautious system instructions to achieve cautious reasoning.
Moreover, Baize \cite{xu2023baize} proposes an open-source chat model that generates a high-quality multi-round dialogue corpus by leveraging ChatGPT to engage in a conversation with itself.
Baize employs questions from Quora3 and Stack Overflow4 as seeds to generate 111.5k dialogues through self-chat.
LongForm \cite{koksal2023longform} generates instructions via LLMs via reverse instructions and builds an instruction-following text dataset, offering a cost-effective and fast approach to perform instruction tuning and output high-quality synthetic data.
To fine-tune instructions to optimize Code Large Language Models (Code LLMs), WizardCoder \cite{luo2024wizardcoder} enhances the fine-tuning of complex instructions for Code LLMs by adapting the Evol-Instruct method to the code domain.
WizardCoder produces intricate code instruction set to improve StarCoder \cite{li2023starcoder} model through code-specific Evol-Instruct \cite{xu2023wizardlm}.

\textbf{Multi-Modality.}
Multi-modality generates cross-modality data via LLMs \cite{dubey2024llama}.
As a typical method, LLaVA \cite{liu2024visual} is the first attempt to extend instruction-tuning to the language-image multimodal domain. It uses ChatGPT and GPT-4 to convert image-text pairs into multimodal instruction-following data and then fine-tunes on the generated instructional vision-language data by combining the visual encoder CLIP \cite{radford2021learning} and language decoder Vicuna \cite{chiang2023vicuna}. 
On this basis, LLaVA-1.5 \cite{liu2024improved}, LLaVA-Plus \cite{liu2023llava}, LLaVA-Interactive \cite{chen2023llava}, and LLaVA-Med \cite{li2024llava} further extend LLaVA to a variety of multimodal tasks and design specialized prompt templates for better fine-tuning.
For example, LLaVA-Plus is dedicated for tool and skill uses in human-AI interaction sessions by incorporating user instructions that request multimodal tools and their execution results into LLaVA.
LLaVA-Med instructs instruction-following data from the captions through GPT-4 to capture open-ended conversational semantics, building a vision-language conversational assistant to answer questions about biomedical images.

\subsubsection{Model Self-Improvement}
Model self-improvement aims at bootstrapping synthetic data from the model itself, including uni-modal synthesis and multi-modal synthesis.

\textbf{Uni-Modality.} This category generates uni-modality data to implement instruction-tuning via the LLM itself.
For example, Self-Instruct \cite{wang2023self} prompts an off-the-shelf GPT3 to generate both new instructions and corresponding instances. It enhances GPT3's ability to follow instructions by leveraging its own generated outputs.
Motivated by backtranslation methods, Instruction Backtranslation \cite{li2023self} generates instructions from human-written “responses” from web documents, instead of generating responses from instructions. It adopts a self-curation step to select high-quality pairs and produce augmented instruction-response pairs.
In addition, SPIN \cite{chen2024self} designs a self-play mechanism like generative adversarial networks (GAN) to achieve instruction tuning. It adopts instances of the same LLMs from different iterations to combine the player (discriminator) and the opponent (generator). 
To beyond human data, ReST$^{EM}$ \cite{singh2023beyond} proposes a two-step self-training method via expectation-maximization for reinforcement learning. It first generates multiple samples for each instruction and filters them to create synthetic data through binary feedback, and then fine-tunes on model-generated synthetic data.
To explore the harmlessness from AI feedback, CAI \cite{bai2022constitutional} generates self-critiques and revised responses via designed principles or instructions (i.e., constitution) and then fine-tunes the original model to achieve self-improvement on harmlessness.
To better teach LLMs to use tools, Toolformer \cite{schick2024toolformer} allows LLMS to automatically transform the original input into the input for API calls via the prompt. In this way, LLMS can teach themselves to use external tools via simple APIs in a self-supervised way.

\textbf{Multi-Modality.}
The above designs various instruction samples to improve the LLM’s
alignment ability. However, these works are usually text-only. Another category synthesizes multi-modality data via the LLM itself.
To fine-tune the large Vision-Language Model (VLM) to improve Visual Question Answering (VQA), SelTDA \cite{khan2023q} designs a self-taught data augmentation strategy for finetuning large VLMs on small-scale VQA datasets.
SelTDA generates question-answer pseudo-labels directly conditioned on an image alone by prompting the BLIP \cite{li2022blip} VLM to caption images, allowing us to pseudo-label unlabeled images.

\subsubsection{Data Augmentation}
Data augmentation aims at enhancing model performance by diversifying training samples without the requirement for extra data. It leverages high-quality instructions or prompts to generate augmented data that users expect and match the target tasks. There are three main types: data labeling, data reformation, and co-annotation.

\textbf{Data Labeling.} Data labeling employs the language comprehension abilities of LLMs to annotate unlabeled examples \cite{zhu2023can}.  
For example, ChatGPT-4 is used to classify political Twitter messages through specific instructions \cite{tornberg2023chatgpt}. Subsequently, some works reveal that these data augmentation ways derived from open-source LLMs are better than manual labeling in many annotating high-resource tasks \cite{zhu2023can,alizadeh2023open}.
An LLM-based prompt strategy evaluates various factors for prompt template and demonstration to augment machine translation data \cite{zhang2023prompting}. It adopts pseudo parallel prompts from monolingual data via zero-shot prompting to improve translation performance.
To achieve multi-modal data augmentation, T-SciQ \cite{wang2024t} teaches science question answering with chain-of-thought (CoT) prompting format to generate high-quality CoT rationales. It is eventually used to train much smaller models to perform CoT reasoning in complex modalities.

\textbf{Data Reformation.}
Data Reformation transforms the existing data into other variations to meet the data format requirements of the target task.
For instance, CORE \cite{dixit2022core} proposes a retrieval-augmented generation approach by creating diverse counterfactual perturbations for counterfactual data augmentation.
It incorporates counterfactual excerpts into prompts to GPT-3 with few-shot learning abilities for counterfactual editing.
Moreover, ALIA \cite{dunlap2023diversify} summarizes the captions of images into short descriptions by prompting GPT-4, and then performs language-guided image editing of the training data with Stable Diffusion \cite{rombach2022high}.
In the NLP task, ChatAug \cite{dai2023chataug} transforms every sentence within the training samples into numerous conceptually akin yet semantically distinct instances.

\textbf{Co-annotation.}
Co-annotation aims to collaboratively annotate data derived from humans and LLMs.
As a representative method, Coannotating designs the human-LLM co-annotation paradigm by using variational prompts to generate responses, which utilizes uncertainty to estimate LLMs’ annotation abilities \cite{li2023coannotating}.
To improve annotation accuracy, ChatGPT is used to augment annotation data with phrases identified as pertinent attributes and brief explanations \cite{bertaglia2023closing}.
To fine-tune code generation models \cite{chen2021evaluating,nijkamp2022conversational} to assist in high-quality code generation, ToolCoder \cite{zhang2023toolcoder} develops an automated data annotation approach for incorporating tool usage information into the source codes through API-augmented prompts.

\subsection{Preference Alignment}
\label{subsec:preference-alignment}
Preference alignment is achieved by systematically refining large models to match complex human preferences \cite{calandriello2024human,gao2024towards}. This process starts with General model distillation \cite{askell2021general,chiang2024chatbot,bai2022training,lee2023rlaif,cui2023ultrafeedback,kopf2024openassistant,wang2023helpsteer,li2024selective,an2023learning,madaan2024self}, which synthesizes broad preference data, providing a foundational alignment across diverse tasks. Domain model distillation \cite{xu2021bot, gehman2020realtoxicityprompts, dai2023safe, ji2024beavertails,stiennon2020learning,luo2024wizardcoder,luo2023wizardmath}  then optimizes models with specialized datasets, enhancing performance in specific domains. Model self-improvement \cite{bai2022constitutional,lee2023rlaif,guo2024direct,ye2024self,yuan2024self,sun2024salmon,hong2023cyclealign,chen2023alpagasus} allows models to iteratively refine their capabilities with minimal human intervention, using self-generated feedback. Data augmentation further strengthens model generalization by expanding and diversifying the training data. These interconnected methods form a coherent framework for optimizing model alignment with both general and domain-specific human preferences.  

\subsubsection{\textbf{General Model Distillation}}

General model distillation aims to generate high-quality preference data by leveraging large language models (LLMs) and external tools to better align models with complex human preferences \cite{askell2021general}. This process is crucial for improving LLM performance in practical applications, particularly in areas like safety, reliability, and ethical considerations \cite{chiang2024chatbot}. One of the primary challenges in this approach is the bias and limitations inherent in strong models \cite{bai2022training}. To address this, distillation from multiple strong models, rather than relying on a single one, can be employed to reduce bias and increase the diversity of responses.

Building upon these strategies, several approaches have been developed to refine preference alignment and mitigate the aforementioned challenges. For instance, RLAIF \cite{lee2023rlaif} synthesizes preference data using sources like Reddit TL;DR~\cite{volske2017tl} for summarization tasks and aligns dialogue generation with human preferences through Anthropic's Helpful and Harmless Human Preferences. Similarly, ULTRAFEEDBACK~\cite{cui2023ultrafeedback} utilizes GPT-4 to generate over a million feedback points. By employing techniques such as best-of-n sampling~\cite{nakano2021webgpt} and Proximal Policy Optimization (PPO)~\cite{lambert2024rewardbench}, it enhances feedback quality and minimizes annotation bias.

In addition to these methods, large-scale datasets have been created to enhance model alignment through crowd-sourced annotations. For example, Open Assistant \cite{kopf2024openassistant} was developed with contributions from over 13,500 volunteers, resulting in more than 161,000 messages across 66,497 conversation trees. Each message is annotated for quality, creativity, and harmfulness. Furthermore, HelpSteer \cite{wang2023helpsteer} enhances data quality by annotating 37,000 conversations for attributes like helpfulness, correctness, coherence, complexity, and verbosity.
Another crucial technique in improving model alignment is Selective Reflection-Tuning \cite{li2024selective}, which refines responses by filtering the teacher model's outputs before using them for distillation. The filtering is based on the student model's r-IFD score, ensuring that only the most challenging and appropriate responses are retained for training. Additionally, models like LEMA \cite{an2023learning} enhance the refinement process by using GPT-4 to identify and correct errors in the student model's responses. These corrections are then used to fine-tune the student model, making the alignment process more accurate and effective.

The refinement and critique capabilities of LLMs are critical to improving alignment. SELF-REFINE \cite{madaan2024self} allows models to critique their own responses, generating improved outputs based on their own feedback. Furthermore, evaluation methods such as MetaCritique from the Critique of Critique \cite{sun2024critiquecritique} provide metrics to assess how effectively a model's critique improves refinement. CriticBench \cite{lin2024criticbench} also explores the relationship between generative capacity and the ability to critique and correct responses, offering insights into model performance.

\subsubsection{\textbf{Domain Model Distillation}}

Domain model distillation focuses on optimizing models for specific tasks by training them on specialized and domain-specific datasets, often using reinforcement learning and preference modeling techniques. This approach enables models to perform well across various domains, enhancing their ability to handle complex, specialized tasks. Through this process, models are refined to meet the requirements of various fields, including Safety-oriented Scenarios \cite{xu2021bot, gehman2020realtoxicityprompts, dai2023safe, ji2024beavertails}, Summarization \cite{stiennon2020learning}, Mathematical Problem Solving \cite{lightman2023let}, Search-based Question Answering \cite{nakano2021webgpt}, as well as Code Generation and Logical Reasoning \cite{luo2024wizardcoder, luo2023wizardmath}.

\textbf{Safety-oriented Scenarios.}
In sensitive or adversarial environments, ensuring safe deployment is essential. BAD \cite{xu2021bot} addresses this by collecting adversarial dialogue data where annotators intentionally provoke unsafe behaviors in chatbots, helping train models to detect and prevent harmful responses. Datasets like REALTOXICITYPROMPTS \cite{gehman2020realtoxicityprompts} with 100K sentence-level prompts annotated for toxicity.  Inspired by Safe RLHF \cite{dai2023safe}, BEAVERTAILS \cite{ji2024beavertails} synthesizes data from over 44,000 red-teaming prompts, generating QA pairs with safety meta-labels to reduce harmful content and improve safety alignment.  

\textbf{Summarization.}
In text summarization, Stiennon et al. \cite{stiennon2020learning} generate high-quality summaries by comparing pairs of summaries from the Reddit TL;DR~\cite{volske2017tl}and CNN/DM datasets~\cite{see2017get}. Human evaluators provide pairwise comparisons for these summaries, which are then used to train a reward model. This model is further fine-tuned using reinforcement learning, refining summarization outputs to align with human preferences for clarity and quality, particularly in domains like news and social media.

\textbf{Mathematical Problem Solving.}
For mathematical reasoning tasks, PRM800K \cite{lightman2023let} offers a large dataset of 800,000 step-level labels across 75,000 solutions to 12,000 math problems. Labelers assign positive, negative, or neutral labels to each step, allowing models to focus on logical consistency and correctness. This approach reinforces the model’s ability to solve complex problems through step-wise reasoning, improving mathematical problem-solving capabilities.

\textbf{Search-based Question Answering.}
For search-based QA, WebGPT \cite{nakano2021webgpt} trains models using long-form question-answering datasets such as ELI5, TriviaQA, and ARC. By interacting with a web-browsing environment, the model generates answers and compares them with human evaluations. This feedback loop improves the model’s search capabilities and performance, particularly in tasks that require sourcing answers from the internet.

\textbf{Code Generation and Logical Reasoning.}
WizardCoder \cite{luo2024wizardcoder} and WizardMath \cite{luo2023wizardmath} have also advanced instruction generation in specific domains like coding and logic-based tasks. By extending the initial WizardLM framework, these models improve the diversity of instructions for code generation and math problem-solving, helping the model handle a wide range of domain-specific challenges.

\subsubsection{\textbf{Model Self-Improvement}}
Model self-improvement focuses on enabling weaker LLMs to iteratively enhance their performance without requiring additional human-annotated data. This approach consists of two categories: self-feedback loops \cite{bai2022constitutional,lee2023rlaif,guo2024direct,ye2024self,yuan2024self,sun2024salmon,hong2023cyclealign}, where the model autonomously refines its outputs based on self-generated feedback, and external evaluation models \cite{chen2023alpagasus,li2023self,dong2023steerlm,yuan2024self,chen2024self}, which rely on external evaluators to assess the model’s responses. Both methods aim to create a scalable system of improvement by reducing dependency on human intervention, allowing the models to continuously optimize their performance through internal adjustments or external guidance.

\textbf{Self-Feedback Loops.}
One of the early methods that exemplify self-improvement through feedback loops is CAI \cite{bai2022constitutional}, which synthesizes alignment datasets by blending human and model-generated prompts using few-shot prompting. By focusing on tasks such as red teaming and helpfulness, CAI enables iterative improvement through AI self-critique and chain-of-thought reasoning, reducing reliance on human feedback. This laid the foundation for RLAIF \cite{lee2023rlaif}, where AI selects preference data that aligns with constitutional requirements. However, early RLAIF faced challenges like distribution shifts due to offline data generation. To address this, methods like OAIF \cite{guo2024direct}and SELF-JUDGE \cite{ye2024self} introduced on-policy preference selection, where a pre-trained Judge model selects preferences in real-time, ensuring alignment with the LLM’s current state.
A key aspect of self-feedback loops is the role of reward models in refining responses. In earlier methods, reward models were often static, leading to problems such as reward hacking~\cite{skalse2022defining}. Self-Rewarding \cite{yuan2024self} introduced a dynamic solution by allowing the LLM to act as its own reward model, iteratively selecting preference data and improving itself through Direct Preference Optimization (DPO)\cite{rafailov2024direct} training. This approach ensures that both the model and the reward mechanism evolve together, maintaining alignment throughout the training process.
Another method for dynamic preference adjustment within self-feedback loops is SALMON \cite{sun2024salmon}, which introduced an Instructable Reward Model that allows for flexible scoring of responses based on different principles. This adaptability enables more precise preference alignment during training. Additionally, CycleAlign \cite{hong2023cyclealign} uses the probability of LLM-generated outputs to rank similar responses, selecting the longest common subsequence of two ranking results to refine the final ordering.

\textbf{External Evaluation Models.}
External evaluation models play an important role in several self-improvement frameworks. For example, ALPAGASUS \cite{chen2023alpagasus} employs a strong LLM like ChatGPT to filter out low-quality data, significantly reducing the training set size while improving model performance. This method demonstrates how external evaluation can enhance model refinement by focusing only on high-quality inputs.
Another prominent technique is Instruction Backtranslation \cite{li2023self}, which generates instruction prompts for unlabeled web data and selects high-quality pairs for fine-tuning. This approach boosts the model’s ability to follow instructions without requiring large-scale human annotations. SteerLM \cite{dong2023steerlm} takes this a step further by fine-tuning models based on specific attributes like humor, creativity, and toxicity. This is done through an attribute prediction model that evaluates responses and refines them using datasets such as Open Assistant \cite{kopf2024openassistant} and HH-RLHF \cite{bai2022training}.

Recent approaches like Self-Rewarding \cite{yuan2024self} and SPIN \cite{chen2024self}  further integrate preference optimization with iterative feedback systems. Self-Rewarding employs the DPO framework, where LLMs generate new prompts and candidate responses, which are then judged by the LLM itself, continuously improving alignment. SPIN, in contrast, eliminates the need for reward models by relying entirely on human-annotated data for iterative improvement. However, as SPIN notes, this approach can become bottlenecked once the model reaches human-level performance, as further improvement requires continuous human intervention.

\subsubsection{\textbf{Data Augmentation}}
Data augmentation is essential for enhancing large model alignment by creating task-specific variations of existing data, which strengthens model generalization and robustness. This approach increases the diversity of training data without the need for additional data collection. Techniques like Data Labeling \cite{zhu2023starling,zeng2024automatic,yuan2024advancing,lin2024criticbench}, Data Reformation \cite{wang2023self,honovich2022unnatural,wu2023lamini,zeng2024automatic,li2023self}, and Co-Annotation \cite{ji2024beavertails,wang2024helpsteer2} are employed to ensure that the augmented data remains relevant and consistent, contributing to more precise model performance across various tasks.  

\textbf{Data Labeling.}
Data labeling plays a crucial role in aligning models with human preferences by providing structured, high-quality feedback to guide learning. Starling-7B \cite{zhu2023starling} is an example of this, collecting ranked data from chat prompts and generating millions of pairwise comparisons to refine model alignment. A more dynamic approach, inspired by Instruction Evolution \cite{zeng2024automatic}, is seen in UltraInteract \cite{yuan2024advancing}, which constructs a Preference Tree from LLM responses. This tree refines incorrect responses based on feedback from models like GPT-4, creating more diverse and robust preference data.   These advancements reflect the need for dynamic evolution in instructions to enhance preference labeling and refinement, as CriticBench \cite{lin2024criticbench} suggests that feedback may be more effective than generation in certain knowledge domains.  

\textbf{Data Reformation.}
 Data reformation is the process of restructuring existing data to better align with task-specific objectives, enabling models to improve their adaptability and performance. A prominent method in this area is in-context learning~\cite{dong2022survey}, where examples embedded in prompts guide LLMs to generate responses that reflect the provided patterns. This closely aligns with Instruction Evolution, which emphasizes increasing the complexity and diversity of instructions. Early works like Self-Instruct \cite{wang2023self} and Unnatural Instructions \cite{honovich2022unnatural}relied on task pools with hand-crafted seed examples, while LaMini-LM \cite{wu2023lamini} expanded this approach by incorporating rich data from Wikipedia to generate more diverse instructions. Auto Evol-Instruct \cite{zeng2024automatic}, initially developed to evolve instructions, automates the optimization of evolution rules by allowing an Optimizer LLM to iteratively improve the rules based on evolving feedback data. Additionally, Instruction Backtranslation \cite{li2023self} enhances instruction-following abilities by generating instruction-response pairs from unannotated data, reducing the need for manual annotation. This ongoing refinement in data reformation is crucial for boosting performance across a variety of tasks.  
 
\textbf{Co-Annotation.}
 Co-Annotation refers to the collaborative process of combining human and machine-generated annotations to improve the quality and safety of model outputs. For instance, BEAVERTAILS \cite{ji2024beavertails} demonstrates how synthesizing data from over 44,000 adversarial prompts, with both human and machine input, generates QA pairs with safety meta-labels, helping models avoid harmful outputs. Similarly, HelpSteer2 \cite{wang2024helpsteer2} blends human expertise with LLM-generated responses to refine multi-turn conversations, ensuring that models adhere more closely to ethical guidelines.

\subsection{Applications}
\label{sec:application}

Most large language models (LLMs) are pretrained and finetuned on general-purpose corpora. To apply them effectively to downstream tasks, continuous task-specific pretraining or finetuning is often required, where data quality and relevance become critical for performance on specialized tasks. However, unlike the abundance of general-purpose data, domain-specific datasets are often scarce due to the intensive knowledge required for their creation. To address the problem, many studies have explored the synthesizing of specialized data with diverse characteristics tailored to each application.

\subsubsection{\textbf{Math}}

Applying LLMs in mathematical scenarios, which involves question understanding and answering, requires intensive logical reasoning. Many researchers have proposed that generating more rationale corpus~\cite{yue2024mammoth, taylor2022galactica, yu2024metamath,zelikman2022star,luo2023wizardmath,xin2024deepseekprover} and diverse questions and answers~\cite{abdullin2024synthetic, liu2024augmenting} in the training corpus helps the model to better understand and reason.

Some studies focus on generating chain of thoughts (CoTs) that explicitly outlines the reasoning steps, either by data augmentation or through LLMs.
Galactia~\cite{taylor2022galactica} proposes a working memory token, wrapping step-by-step reasoning within ``<work> </work>'' for 4 datasets to form a new mathematical reasoning dataset. Additionally, recent literature harnesses the reasoning capability of advanced closed-source LLMs. MammoTH~\cite{yue2024mammoth} compiles 13 math datasets with intermediate rationales to form the MathInstruct dataset, and extends them with hybrid CoT and PoT (Programe-of-Thought) rationales with the help of GPT-4.

Diverse questions and solutions can also enhance the math understanding and reasoning capability of LLMs. General purpose LLMs like GPT-4 are utilized to model linear programming of a specific real-world problem~\cite{abdullin2024synthetic} through conversations between two LLM agents, and supervised by another GPT-4 agent for quality and correctness. MetaMath~\cite{yu2024metamath} increases the diversity of questions by rewriting mathematical questions from two reasoning paths and rephrasing using LLMs. Some methods attempt to compose questions from seed questions. For example, Liu et al.\cite{liu2024augmenting} propose an Iterative Question Composing (IQC) method, which uses an LLM to compose questions from seed questions and employs another LLM to conduct rejection sampling.

As mathematical questions and answers are verifiable, some approaches expand the training corpus through self-generated formulated problems and proofs that have been verified by themselves or external tools and models.
STaR~\cite{zelikman2022star} generates rationales to answer many answers with few rationale examples as a prompt to bootstrap the reasoning capability of GPT-J~\cite{wangKingoflolzMeshtransformerjax2024}. The answers are verified and, if the generated answers are wrong, the model will retry until the rationale gives the correct answer. All correct rationales are used for fine-tuning.
DeepSeekProver~\cite{xin2024deepseekprover} is initialized using DeepSeekMath-Base~\cite{shao2024deepseekmath} finetuned on MMA dataset~\cite{jiang2023multilingual} to acquire basic auto-formalization capability. The model generates formulated math problems and proofs, and filters the generated data with miniF2F-valid as few-shot examples. The proofs are generated for both original theorems and their negations as an augmentation approach. All valid generated data will be used to train the model, and the process is repeated several times until a marginal performance gain is observed.
WizardMath~\cite{luo2023wizardmath} generates evolved instructions using ChatGPT and also provides supervision during the generation process. It also evolves the Wizard-E model through Reinforcement Learning from Evol-Instruct Feedback (RLEIF), alongside Evol-Instruct and Reinforcement Learning, to improve LLM reasoning performance.

\subsubsection{\textbf{Science}}

Scientific applications require a deep understanding of knowledge-intensive concepts and reasoning, which requires high-quality data sets for effective instruction fine-tuning. However, generating such data sets is challenging due to the varying formats across disciplines and the underlying logic can be difficult to articulate.

Unifying the format of different disciplines is the first step in processing science-related corpus, by transforming structured data into readable texts~\cite{li2024scilitllm, zhang2024chemllm,zhang2024sciglm} or a specialized tokenizer~\cite{taylor2022galactica}.

The instruction tuning data sets are then generated from the collected raw data. To ensure diversity, SciLitLLM~\cite{li2024scilitllm} proposes a three-step pipeline to generate diverse and high-quality instructions for scientific contexts. It first collects keywords and task descriptions for different scientific domains from domain literature and datasets, and then prompts GPT-4o with keywords and descriptions sampled from previously collected data. 
ChemLLM~\cite{zhang2024chemllm} employ a Play as Playwrights CoT style while prompting GPT-4 to promote context richness and logical coherence.
Chain-of-thought also aid the model to better understand scientific topics.
SciGLM~\cite{zhang2024sciglm} collects the raw questions on college-level physics, chemistry, math, and formal proofs, generating CoT for them using GPT-4 with a self-correction~\cite{shinn2023reflexion, li2023reflectiontuning} process. An instruction quality classifier is trained with positive data labeled by LLMs and humans, and negative data identified by LLMs.
Galactia~\cite{taylor2022galactica} focuses on knowledge-intensive scientific tasks: equations, chemical reactions, and citation prediction, and proposes a set of tokenizers for various input formats. It also creates a working memory technique for better in-context reasoning.

\subsubsection{\textbf{Code}}

Generating synthetic data that enhance coding performance has long been studied, which requires a clear understanding of questions and accurate reasoning to produce correct code. Since the accuracy of the code can be easily validated in a simulated coding environment, this enables the generation of large-scale instruction tuning datasets for coding tasks.
Haluptzok et. al.~\cite{haluptzok2023language} proposed a self-play method to generate programming puzzles~\cite{haluptzok2022generating} and its solution, where the correctness is verified by a Python interpreter. They further finetune the LLM on the generated data to improve the performance.
For more time-efficient code generation, Shypula et. al.~\cite{shypula2024learning} present a data generation method for code optimization through self-play and few-shot Chain-of-Thoughts. They prompt GPT-3.5 to generate more optimization problems on the proposed PIE dataset, and then the answer is generated by a finetuned GPT-3.5 on the PIE dataset, resulting in more samples.
WizardCoder~\cite{luo2024wizardcoder} utilizes StarCoder 15B as the foundation and finetunes it using the code instruction-following evolved through Evol-Instruct~\cite{xu2024wizardlm}.
Code Alpaca~\cite{chaudhary2023code} is fine-tuned from a 7B and 13B LLaMA model on 20K instruction-following data generated by the techniques in the Self-Instruct~\cite{wang2023selfconsistency}.

Diverse programming problems and the corresponding answers promote the capabilities of LLMs.
MagicCoder~\cite{wei2024magicoder} utilizes open-source code snippets to generate more diverse, realistic, and controllable coding instruction data.
Code LLama~\cite{roziereCodeLlamaOpen2024} constructs a dataset of 14,000 question-test-solution triplets by generating 52,000 unique programming questions with Llama 2 70B and using Code Llama 7B to create unit tests and Python solutions.
Phi-1~\cite{gunasekar2023textbooks} enhances diversity by constraining topics and target audiences when generating text-code interleaved textbooks using GPT-3.5 as a training corpus. Its successor, Phi-1.5~\cite{li2023textbooks}, expands this approach by creating synthetic textbook-style data for common sense reasoning and world knowledge, while incorporating samples from web datasets to further increase diversity.

\subsubsection{\textbf{Medical}}

In medical applications, large models mainly serve as medical dialogue chatbots, and need to interact with patients with multi-round dialogues. To achieve interactive data synthesis, specialized documents are first collected as seed corpus, such as medical diagnostic records. Based on that, diverse question-and-answer pairs can be generated with the aid of general-purpose large language models, and used to improve understanding ability to produce helpful responses.

To simulate conversations between the patient and the doctor, DISC-MedLLM~\cite{bao2023discmedllm} refines the MedDialog~\cite{zeng2020meddialog} and cMedQA2~\cite{zhang2018multiscale} datasets by leveraging GPT-3.5 to generate high-quality conversation samples. This process involves the removal of colloquial expressions, resolution of inconsistencies, and standardization of language for greater uniformity. 
HuatuoGPT~\cite{zhang2023huatuogpt} utilizes data distilled from ChatGPT through self-instruct~\cite{wang2023selfconsistency} and conversations between two ChatGPTs.
HuatuoGPT-II~\cite{chen2023huatuogptii} introduces a one-stage adaptation strategy that merges traditional continued pre-training and supervised fine-tuning into a single step of supervised fine-tuning. It employs GPT-4 for data unification, converting domain-specific pre-training text into a (instruction, output) format with a standardized language and style.
ChatCounselor~\cite{liu2023chatcounselor} is a mental health support trained on GPT-4 refined query-answer pairs based on doctor-patient dialogues, where GPT-4 also generates summaries of key information from each interaction for broader context. 

Privacy is also a primary issue when dealing with sensitive medical records, some literatures extract medical knowledge from knowledge graphs, and generate synthetic medical text without any personal information. ClinGen~\cite{xu2024knowledgeinfused} alleviate this by generating synthetic clinical text from Knowledge Graphs~\cite{su2023biomedical} and chatGPT. DISC-MedLLM~\cite{bao2023discmedllm} synthesizes medical dialogues by incorporating domain-specific knowledge graphs question-answer pairs.


\subsubsection{\textbf{Law}}

LLM-based legal assistants have gained considerable attention for providing affordable and convenient legal services, particularly in the areas of legal question answering and consultation. Recent studies focus on the quantity and quality of finetuning datasets by using data synthesis to improve the clarity and formality of the responses.
DISC-LawLLM~\cite{yue2023disclawllm} design prompts for GPT-3.5-turbo to refine the legal syllogism, knowledge expansion, and Chain-of-Thought reasoning for the SFT instruction dataset. 
LawyerLLaMA~\cite{huang2023lawyer} uses ChatGPT to generate explanations for each question, or to generate explanations based on questions and answers. In this work, the data generated by human experts produces better results than only 6k and 42k automatically generated data. They identify that more automatically generated data may achieve better performance, and mark it as future work.
LawGPT~\cite{zhou2024lawgpt} refine open-source datasets by prompting chatGPT for instruction-tuning data to generate more formal, polite, and clear answers.
WisdomInterrogatory~\cite{zhihaillm2024zhihaillm} prompt GPT-3.5 models as agents to imitate conversations between a law agent and a user to generate multi-turn instruction text.

\subsubsection{\textbf{Others}}

In addition to these previous applications, the potential of synthetic datasets is also explored in financial~\cite{bhatiaFinTralFamilyGPT42024} and education~\cite{doughty2024comparative, latif2024knowledge}. These areas are more challenging for data synthesis due to the higher knowledge density and demands on quality. As research continues, these areas may become increasingly promising.



\section{Functionality}
\label{sec:function}
From the functional perspective of LLMs, data synthesis and augmentation can be divided into four categories: understanding, logic, memory, and generation \cite{touvron2023llama,chowdhery2023palm}. 
By exploring the four basic functions in LLMs, data synthesis and augmentation can fully capture the inherent patterns in large-scale data and effectively apply them to downstream applications and tasks.

\subsection{Understanding}
\label{sec:Understanding}
Understanding functionality leverages the powerful language understanding of LLM to understand the inherent patterns in the data.
From the perspective of the content of understanding, it includes uni-modal and multi-modal understandings. Uni-modal understanding mainly comprehends the semantics of texts, including text understanding and semantic labeling. Multi-modal understanding combines multiple modalities.

\textbf{Text Understanding.} 
To improve the instruction-following capacities of LLMs, Alpaca \cite{Taori2023alpaca} and Alpagasus \cite{chen2023alpagasus} utilize human-written instruction-response pairs to generate instruction-following demonstrations via few-shot prompting, improving the capabilities of instruction-following models.
Along this line, WizardLM \cite{xu2023wizardlm} generates more diversified and multitask-oriented instructions from both depth and breadth to improve the performance of LLMs.
To perform the model self-improvement, Self-Instruct seeks to improve the instruction-following capabilities of LLMs by bootstrapping off their own generations \cite{wang2023self}. It classifies the generated instruction and generates possible class labels to promote diversity.
To further comprehend text content, SPIN \cite{chen2024self} considers instruction-tuning as a two-player game by using the main player as the discriminator to distinguish the responses between LLM and humans. This strengthens LLM without the need for additional human-annotated data by understanding and discriminating textual responses.
Moreover, WRAP\cite{maini2024rephrasing} capitalizes on the inherent diversity of web articles, enabling LLMs to produce high-quality paraphrases of noisy and unstructured content found online.

\textbf{Semantic Labeling.} 
For a deeper understanding of text properties, Instruction Backtranslation promotes instruction following the language model by automatically labelling human-written documents with corresponding instructions \cite{li2023self}.
Analogously, ChatGPT-4 is used to classify political messages through instructions by labeling them with political tendencies \cite{tornberg2023chatgpt}, further understanding semantics from messages.
Introducing the data augmentation pattern of ChatGPT annotating, Co-Annotating \cite{li2023coannotating} collaboratively views the relation between humans and LLMs and the selected Pareto efficiency concept enables participators to compare strategies and gain a deeper comprehension of tradeoffs between cost and performance.
To offer annotators with explanations, ChatGPT is introduced to improve human labeling of sponsored content on social media by designing prompts to generate model explanations \cite{zhang2023toolcoder}.

\textbf{Multi-Modal Understanding.} 
LLaVA \cite{liu2024visual} is the first work to use LLMs to generate high-quality language-image instruction-following data for general-purpose visual and language understanding. It employs captions and bounding boxes of images as symbolic representations to encode the images as an LLM-recognizable sequence and produces three types of instruction-following data to fully capture semantic information from the images.
It has been demonstrated that synthetic data improves the performance of models for image recognition tasks in data-scarce settings\cite{he2022synthetic}.
ChartLlama\cite{han2023chartllama} is trained on the dataset generated through a multi-step process, encompassing a broader variety of chart and task types than those found in existing datasets.
Genixer\cite{zhao2023genixer} flexibly produces a variety of visual instruction tuning data from unlabeled images, enhancing capabilities for visual question answering and referring expression comprehension tasks.
The quality of SciLitInsis\cite{li2024scilitllm} has been evaluated from five aspects including clarity, complexity, correctness, usefulness, and adaptability to improve LLMs’ ability in scientific literature understanding.

\subsection{Logic}
\label{sec:Logic}
Logic functionality fully taps the reasoning and logic functions in the process of synthesizing and augmenting data. According to the application of logic, there are the following three categories: code logic, math logic, and reasoning.

\textbf{Code.} 
To provide complex reasoning capabilities, ReST$^{EM}$ \cite{singh2023beyond} selects training samples with a binary reward function to output synthetic data. It leverages logical reasoning to improve the performance of LLMs on problem-solving tasks, such as code generation and mathematical problems.
To optimize code, ToolCoder \cite{zhang2023toolcoder} utilizes prompts to generate API-augmented codes for the API search tool with ChatGPT.
The code generation process is enhanced by integrating API search tools, enabling the model to automatically utilize the search tool for suggestions when selecting an API.
Moreover, OSS-Instruct\cite{wei2024magicoder} enlightens LLMs to generate more diverse, realistic, and controllable coding instruction data, which can substantially boost the reasoning performance of various LLMs. 
High-quality Case2Code\cite{shao2024case2code} dataset is automatically and efficiently harvested from pre-training code texts by leveraging small LLMs and a coder interpreter
The LLM’s performance is enhanced through fine-tuning on synthetic programming and solutions which are verified by a Python interpreter\cite{haluptzok2022language}.

\textbf{Math.} 
MathInstruct\cite{yue2023mammoth} has two main characteristics including broad coverage of different math fields and complexity levels, and hybrid CoT and PoT rationales, enabling MAmmoTH to outperform existing open-source models. 
Benefiting from the diverse data sources of MMIQC\cite{liu2024augmenting} and the effectiveness of iterative question composing, the models fine-tuned on MMIQC achieve new SOTAs on math benchmark.

\textbf{Reasoning.} 
For reliable reasoning performance, Orca \cite{mukherjee2023orca} proposes progressive learning from complex explanation Traces of GPT-4 and learns to imitate the reasoning process of LLM.
Further, Orca2 \cite{mitra2023orca} designs various reasoning strategies, e.g., step-by-step and recall-reason-generate, to enhance the reasoning abilities of smaller LLMs. 
To generate synthetic data for instruction-tuning, CAI \cite{bai2022constitutional} sample a critique request from a short list of constitutions and prompt the model to generate a critique of the response, identifying harmful outputs to build a harmless but nonevasive AI assistant \cite{rosati2024representation}.
To enhance the science question answering task, T-SciQ \cite{wang2024t} uses the CoT reasoning capabilities of LLMs to generate QA-CoT samples as teaching data samples. Since some questions are exceptionally intricate, T-SciQ further designs a 3-step zero-shot prompting way to produce planning-based CoT rationales, which breaks down complex problems into simpler subproblems that can be easily solved.
Utilizing CoT examples as a guide, the LLM generates high-confidence CoT reasoning paths which serve as the final training samples to be fed back to the model for fine-tuning\cite{huang2022large}.
In addition, STaR\cite{zelikman2022star} improves the LLM itself by augmenting a fine-tuning dataset using rationalization and justifying ground-truth answers to problems the model failed to solve.
Symbol tuning\cite{wei2023symbol} leverages the intuition that when a model cannot use instructions or natural language labels to figure out a task, it must instead reason with input-label mappings in-context.

Moreover, reasoning can also be extended to multi-modal scenarios.
To optimize the visual question answering task, SelTDA \cite{khan2023q} prompts VLM to caption various images and then converts each caption into a boolean question. By this means, SelTDA builds a direct image-to-text task and pseudo-labels unlabeled images by sampling questions and answers.
Recognizing that LLMs struggle with abstract image perception and reasoning, multimodal self-instruct\cite{zhang2024multimodal} creates a benchmark of abstract images tailored for everyday contexts.

\subsection{Memory}
\label{sec:Memory}
Memory functionality remembers and utilizes previously learned information in LLM when synthesizing data \cite{huang2024agents}. According to the properties of memory content, memory functionality can be divided into three categories: procedural memory, semantic memory, and episodic memory \cite{tulving1985memory}.

\textbf{Procedural Memory.} 
Procedural memory preserves a specific process of how tasks and actions are executed.
To perform the code sorting task for software engineering, a sorting algorithm is used to sort out the code LLMs by recognizing affiliation categories, including different communities and contributors \cite{zheng2308towards}.
To enable the LLM agent \cite{zhao2024expel} to operate automatically, a code generation model employs three types of in-context supervision to specify library functions and construct the agent’s codes for solving user-instructed tasks \cite{patel2023evaluating}.
Additionally, to explore the internal reasoning processes of LLM, Quiet-STaR \cite{zelikman2024quiet} generates thoughts and rationales to explain future text when following all tokens in the text based on Self-Taught Reasoner \cite{zelikman2022star}. 
It resorts to procedural reasoning to produce the process of thinking like humans.

\textbf{Semantic Memory.} 
Semantic memory synthesizes symbolically representable data to retain acknowledged world knowledge like knowledge graphs, documents, and APIs. 
To achieve autonomous knowledge graph construction and reasoning tasks, AutoKG designs a multi-agent-based approach that combines multiple intelligent agents to play different roles like assistant and domain expert to collaborate to complete tasks \cite{zhu2024llms}. Meanwhile, AutoKG incorporates the external knowledge base and Internet resources into the knowledge graph to make up for the limitations of LLMs.
To explore semantic content from general documents, KPC designs a knowledge-driven prompt chaining-based code generation model \cite{ren2023misuse}. It disaggregates the process of code generation into iterative validation and revision phases and employs fine-grained exception-handling knowledge derived from documentation to facilitate LLMs in code generation. 
To effectively utilize existing APIs and prevent the model from inventing non-existent APIs, De-Hallucinator autonomously recognizes API references pertinent to specific projects with the model's preliminary predictions, subsequently incorporating these references into the prompt \cite{eghbali2024hallucinator}.

\textbf{Episodic Memory.} 
Episodic memory remembers contextual content that is closely related to the current state and personal experiences.
To create diverse synthetic data to showcase diverse personas, Persona Hub proposes a persona-driven data synthesis methodology to capture knowledge and experiences from real-world users \cite{chan2024scaling}.
Persona Hub generates 1 billion diverse personas from massive web data by text-to-persona and persona-to-persona ways.
To obtain an accurate response, AceCoder proposes a prompt-enhanced technique to capture context information. It first tells LLM to analyze and clarify requirements and then selects similar programs as supplementary examples in prompts to provide relevant content for prompts \cite{li2023acecoder}.
To effectively utilize user experience information, RepoCoder optimizes the code completion process at the repository level by integrating a similarity-based retrieval mechanism and a pre-trained code language model within an iterative retrieval-generation framework \cite{zhang2023repocoder}.
It adopts available retrievers to locate pertinent knowledge within a repository, thereby improving the contextual foundation for LLM.

\subsection{Generation}
\label{sec:Generation}
Generation functionality aims at producing coherent and contextually relevant content for downstream tasks and applications.
Based on the form of generated content, there are the following two categories: content generation (e.g.,text and multi-modal generation) and retrieval-augmented generation.

\textbf{Text Generation.} 
To accomplish sequence understanding and generation for machine translation, a LLM-based prompt strategy \cite{zhang2023prompting} leverages back-/forward-translation to augment monolingual data via zero-shot prompting.
For text augmentation, ChatAug rephrases a sentence into multiple semantically similar sentences \cite{dai2023chataug}.
Genie \cite{yehudai2024genie} generates high-quality data that is natural, faithful and diverse, and the models trained on the synthetic data are more faithful to the grounding content.
From the guidance of principles of diversity and quality, UltraMedical \cite{zhang2024ultramedical} dataset can be used to build specialized generalists in biomedicine.

By utilizing paraphrase proximity and critic-guided distillation, Impossible Distillation \cite{jung2024impossible} produces a high-quality dataset to enhance the model’s capability of paraphrase generation and sentence summarization.
HuaTuo\cite{wang2023huatuo} enhances the factual accuracy of its responses by initially drawing knowledge instances from a knowledge graph, followed by the creation of instances tailored to that specific knowledge using ChatGPT.
TinyStories\cite{eldan2023tinystories} is designed to capture the essence of natural language, thereby enabling the models trained upon it to produce fluent and consistent stories.
UltraChat is created using two distinct ChatGPT Turbo APIs to craft informative and realistic multi-turn dialogues. One simulates the user to produce queries while the other crafts the response\cite{ding2023enhancing}.

Additionally, Baize\cite{xu2023baize} employs self-distillation techniques that incorporate feedback, enabling the model to discern nuance within the feedback and carry out fine-grained optimization.
DIALOGIC\cite{li2022controllable} efficiently scales a limited dialogue dataset with minimal to zero human involvement and parameter adjustments, addressing the challenge of low-resource scenarios.
DISC-MedLLM\cite{bao2023discmedllm} leverages ChatGPT to rephrase existing medical NLP datasets to provide accurate and truthful medical responses in end-to-end conversational healthcare services.
In ReST\cite{gulcehre2023reinforced}, multiple steps of negative log-likelihood training with progressively increasing filtering thresholds in the improve step and the human evaluation scores in the grow step lead to continuous improvements in the model’s performance.
Self-translate-train\cite{ri2024self} enhances cross-lingual performance by effectively harnessing the model’s inherent translation capabilities.
TinyDialogues\cite{feng2024child} discovered that a diverse blend of data sources outperforms homogeneous conversation data, indicating that high-quality synthetic conversation data yields better performance than natural conversation data.

\textbf{Multi-Modal Generation.} 
To achieve automated language-guided image augmentation, ALIA \cite{dunlap2023diversify} summarizes the image captions to generate natural language descriptions via LLM and shows
significantly visual diversity compared to the original data.

\textbf{Retrieval-Augmented Generation.} 
Retrieval-Augmented Generation (RAG) integrates external knowledge to generate accurate and contextually appropriate content \cite{chen2024benchmarking}.
To address holistic questions across an entire text corpus, GraphRAG \cite{Microsoft2024GraphRAG} utilizes LLMs to construct a graph-based text index and integrates all partial responses to generate a global response. The constructed knowledge graph contains a wealth of entities and relations in GraphRAG, which is conducive to generating more comprehensive and diversified answers.
Considering the limitations of knowledge embedding model and
independent processing between query and context, RankRAG \cite{Yu2024RankRAG} instruction-tunes LLMs to concurrently assess the relevance between queries and contexts and utilizes the retrieved context to generate accurate answers.
To mitigate the imbalance between the burden of
the retriever, LongRAG ~\cite{Jiang2024LongRAG} designs the long retriever and reader components to reduce the corpus size, thereby achieving effective retrieval recall with only a few top units.

\section{Challenges and Limitations}
\label{sec:challenge}
Data synthesis and augmentation play a pivotal role in enhancing the capabilities of LLMs. However, these techniques face significant challenges and limitations that can impede their effectiveness. 

\subsection{Synthesizing and Augmenting Method}
\label{sec:Synthesizing and Augmenting Method}
Despite the significance of synthesizing and augmenting data, there are critical challenges of using different synthesizing and augmenting method in practice. 

\textbf{Dependence on LLMs.}  
The ability of LLMs to perform few-shot or zero-shot tasks is leveraeged to generate data, suggesting that these models need to be sufficiently large to possess a certain level of reasoning or data generation capabilities \cite{yang2024synthesizing,lee2024llm2llm}. Otherwise, data synthesis or augmentation with smaller models may not be sufficient to enhance teh specific capabilities of LLMs. Hence, the methods of data synthesis and augmentation seriously rely on the abilities of LLMs themselves.

\textbf{Complicating Evaluation and Depollution in Model Training}. 
The use of synthetic data in model training can significantly complicate fair evaluation. Evaluation benchmarks are often created based on public text sources (such as course websites or forums), and the introduction of synthetic data can exacerbate this issue. While the community has proposed several techniques for detecting evaluation pollution, these token-based depollution methods may prove ineffective when synthetic data is involved in model training \cite{golchin2023time}. It is recommended that model developers invest in creating and maintaining internal and protected evaluation benchmarks to ensure the integrity of the evaluation process.

 \textbf{Uncertainty and Search Complexity in RLAIF.} Preference alignment has progressed from human feedback to AI-generated feedback. Despite these advancements, current methodologies remain largely dependent on passive feedback mechanisms, where meaningful insights are only gathered through active evaluation and querying of the model's performance \cite{lee2023rlaif}. This poses a significant challenge, as the expansive search space demands a considerable number of test samples to acquire valuable feedback. Moreover, a critical concern with AI-generated feedback is the uncertainty regarding its true alignment with human preferences. To date, no definitive evidence has been presented to confirm that AI feedback can reliably function as an effective alignment signal, raising doubts about its credibility and validity.

\textbf{Unstable and Inconsistent Logical Paths.} In applications that depend significantly on logical reasoning, researchers strive to generate synthetic reasoning paths~\cite{zelikman2022star, luo2023wizardmath, zhang2024chemllm, zhang2024sciglm, shypula2024learning} to enhance the logical capabilities of large language models (LLMs), often verifying only a subset of the steps or only the answers. However, verifying the correctness and consistency of every step in these reasoning paths remains challenging, particularly in knowledge-intensive disciplines. Integrating knowledge bases or formal logical reasoning tools into data generation systems may facilitate the creation of more stable and consistent logical paths.

\subsection{Data Quality}
\label{sec:Data Quality}
Data quality is paramount particularly when it comes to synthetic and augmented data for LLMs. Unlike the diverse, credible, high-quality real data that already exists, the nature of data synthesis and augmentation may influence the quality the generated data.

\textbf{Data Diversity.}
Many current methods rely on predefined templates or algorithms that may inadvertently produce homogeneous outputs. This homogeneity restricts the model's exposure to diverse linguistic patterns and contexts, ultimately hindering its ability to generalize to varied real-world inputs \cite{chung2023increasing,cegin2024effects}.
In addition, synthetic and augmented data often fails to capture the intricate nuances of real-world language use, such as idiomatic expressions, cultural references, and contextual variations. This shortcoming arises because synthetic and augmented data may not incorporate the richness and diversity found in authentic conversations and interactions. Consequently, models trained predominantly on these data may struggle with the subtleties of language that are crucial for effective communication and comprehension \cite{nagireddy2024dare}.

\textbf{Long-tail Phenomenon.} The most significant improvements in data generated by LLMs are observed in languages that are widely used. Conversely, there may be only slight advancements in contexts where languages are less frequently used. As a consequence, the methods for data synthesis and augmentation might struggle to effectively handle rare or innovative instructions \cite{wu2024coral,zhao2024ltgc}.   
This causes the long-tail problem that the distribution of language usage is highly skewed. Common phrases and contexts may dominate the synthetic and augmented data, while rare or specialized cases receive little to no representation. This can lead to significant performance drops in scenarios involving infrequent language use, such as technical jargon or niche topics.

\textbf{Reliability}. The reliability of synthetic data cannot be effectively guaranteed \cite{long2024llms,giglou2024llms4synthesis}. For example, despite conforming to linguistic rules, is fluent, and is aligned with human preferences, there are still many problems in the fact and timeliness of the synthetic content, and it is not possible to make a reliable evaluation of the synthesized content. More important, synthesized data can reflect biases or inaccuracies from the original datasets, leading to unreliable outputs \cite{mondal2024mitigating}. If a dataset used for training contains biased samples, the synthesized data will likely perpetuate those biases.

\textbf{Distribution Inconsistency}. 
The inconsistency between the distributions of synthetic and human-generated data is crucial for LLMs. When there is a significant mismatch, models may struggle to understand and generate responses that resonate with users \cite{xiong2023examining}.
First, biases in the original data and LLMs may generate inappropriate data that do not reflect the natural distribution of language usage, leading to distribution discrepancies.
Second, the lack of context in synthetic and augmented data causes the generated phrases or sentences to be out of alignment with the pragmatics of human communication.

\subsection{Impact of Data Synthesis and Augmentation}
\label{sec:Impact of Data Synthesis and Augmentation}
In addition to the quality and methodological factors of the data itself, data synthesis and augmentation can also have a certain impact on the external environment, involving individuals, groups, societies, and human values.

\textbf{Privacy}.
Synthetic and augmented data may contain traces or patterns derived from real individuals, thereby compromising user privacy. This challenge is exacerbated in applications where the generated data mimics sensitive attributes, making it difficult to ensure complete anonymity \cite{yan2024protecting,wang2024harmonic}.    
Moreover, to address concerns over ownership and authenticity, data watermarking techniques are being explored. These techniques allow for the embedding of identifiable information within synthetic data, enabling traceability and accountability. However, the implementation of such systems must balance the need for privacy with the ability to verify data provenance, raising questions about the feasibility of effective watermarking without compromising data utility.

\textbf{Security.}  
While synthetic and augmented data is often seen as a solution to privacy concerns, its security is not guaranteed. The generation process can be exploited by malicious actors to craft adversarial examples that deceive LLMs \cite{zhang2024llms}. Moreover, these generated data can sometimes be reverse-engineered, revealing patterns that can lead to the reconstruction of sensitive real-world data.
Another potential risk is that synthetic and augmented data can also serve as a vector for introducing vulnerabilities into models. Attackers may embed backdoors into the data generation process, leading to models that behave unpredictably under specific conditions. Such vulnerabilities can be exploited for nefarious purposes, raising significant security concerns \cite{wu2024new}.  

\textbf{Social Impacts.}  
Data synthetic and augmentation introduce a myriad of legal challenges. For instance, the authenticity of generated data can complicate issues of intellectual property rights and data ownership \cite{patel2024datadreamer}. 
Furthermore, the misuse of data poses a significant threat to the dissemination of misleading information. In particular, its application in mimicking real individuals, and manipulating public opinion, carries grave risks.
An important fact is that these data can inadvertently perpetuate or amplify societal biases. If LLMs are trained on biased data, the outputs derived from LLMs may reflect and reinforce harmful stereotypes. 
Moreover, cultural norms vary significantly across societies; thus, synthetic and augmented data that is appropriate in one context may be offensive or inappropriate in another. The socio-cultural implications of data should be carefully considered to avoid entrenching existing inequalities \cite{shen2023shaping}.

\subsection{Impact on Different Applications and Tasks}
\label{sec:Impact on different applications and tasks}
Data synthesis and augmentation techniques play a crucial role in enhancing the performance of LLMs across diverse tasks. However, the effectiveness of these methods can vary significantly depending on their implementation and the specific characteristics of the synthetic and augmented data.

\textbf{Generalization}.  
While synthetic and augmented data can be useful for training models in scenarios where labeled data is scarce, its ability to generalize to unseen examples often falls short. This limitation arises from the fact that the generated data may not fully capture the complexity and variability present in real-world data. As a result, models trained predominantly on these data may exhibit poor performance when applied to new, real-world situations, leading to a phenomenon known as overfitting to the specific patterns of data. For instance, in the medical domain, data synthesized from one healthcare system might not be applicable to another due to differing protocols or patient demographics \cite{yang2024unveiling,yang2024exploring}.

\textbf{Transferability and Domain Adaptation}. 
 Synthetic and augmented data can be designed to mimic various domains. However, the transfer performance to different domains remains a challenge. Data pattern from one domain may struggle to adapt to another due to discrepancies between the synthetic and real data distributions \cite{liu2024panda,wang2024role}. Therefore, when data is generated, it often embodies specific characteristics of the source domain, which may not fully align with the target domain. This misalignment can hinder the model's ability to effectively transfer its knowledge. For instance, a model trained on synthetic dialogues that simulate formal language may perform poorly when faced with informal conversational contexts in real-world applications.

\subsection{Future Directions}
\label{sec:Future Directions}
\textbf{Multi-Modal Synthesis}. Multi-modal data synthesis technology addresses the integration of diverse data types (e.g., text, images, audio, and sensor data) to create richer and more informative data \cite{jiang2023cola}. This helps to create diverse training samples that enhance model robustness against over-fitting and variability. These diverse samples effectively alleviate the data scarcity problem in many specific scenarios like healthcare and autonomous systems.

\textbf{Real-time Synthesis}. Advances in real-time data synthesis will allow for dynamic generations, enabling applications like virtual assistants and interactive gaming environments to adapt seamlessly to user inputs \cite{karunya2023ai}. This capability will enhance user engagement and create more personalized experiences across various platforms.

\textbf{Domain Model Distillation.} Existing investigations mainly focus on synthesizing data using a general model, such as GPT. However, domain-specific models have stronger capabilities within their respective domains compared to general models. Therefore, leveraging domain-specific models to synthesize domain data is expected to further enhance the capabilities of LLMs within that domain \cite{lightman2023let,nakano2021webgpt,luo2024wizardcoder}.

\textbf{Large-Scale Synthesis.} Pre-training a large model requires a vast amount of high-quality data. Existing data synthesis or augmentation methods have limited scalability in generating data. The synthesis of large-scale pre-training datasets is expected to enhance the comprehensive capabilities of pre-trained models \cite{karunya2023ai}.

\textbf{Robust Quality Evaluation Metrics.} 
Data synthesis and augmentation need to ensure the quality of generated data \cite{chundawat2022universal}. Current methods lack standardized metrics that can adequately evaluate the diversity and relevance of synthesized datasets. Future research should focus on developing robust metrics that assess not just the linguistic quality of generated text but also its contextual appropriateness, diversity, and potential biases.

\textbf{Ethical Considerations and Responsible Data Synthesis and Augmentation.} 
Ethical considerations and responsible practices in data synthesis and augmentation for LLM are critical to ensuring the integrity and fairness of AI systems. As these methods increasingly influence model training and outcomes, some key issues such as bias propagation, misinformation, and data misuse become obvious. It is essential to establish ethical guidelines that govern the generation and application of synthetic and augmented data, promoting transparency, accountability, and inclusivity \cite{jiao2024navigating}.

\section{Conclusion}

Data synthesis and augmentation are essential for advancing LLMs, particularly in meeting the need for large-scale and high-quality data for LLMs. This survey provides a comprehensive review of LLM-oriented data synthesis and augmentation techniques, systematically exploring their applications across the entire lifecycle and core functions of LLMs, while building a framework that connects existing research, highlights key methods, and clarifies strengths and limitations. We envision that advancements in LLM-oriented data synthesis and augmentation methods will unlock new possibilities to enhance data efficiency, improve generalization across tasks, and drive the evolution of data-centric AI. We hope this survey serves as a foundation for future research, inspiring innovation and progress in the field of LLM-oriented data synthesis and augmentation.

\section*{Acknowledgement}
This research was supported by “Pioneer” and “Leading Goose” R\&D Program of Zhejiang (No. 2024C01020), National Key R\&D Program of China (No. 2023YFF0725600), National Science Foundation of China (No. 62406015).

\bibliographystyle{ACM-Reference-Format}
\bibliography{sample-base}

\end{document}